\documentclass[conference]{IEEEtran}
\IEEEoverridecommandlockouts
\usepackage{cite}
\usepackage{amsmath,amssymb,amsfonts}
\usepackage{algorithmic}
\usepackage{graphicx}
\usepackage{textcomp}
\usepackage{xcolor}
\usepackage{caption}
\usepackage{multirow}
\usepackage{booktabs}
\usepackage{subcaption}
\usepackage[linesnumbered,ruled,vlined]{algorithm2e}
\usepackage{hyperref}  % or \usepackage{url}

\def\BibTeX{{\rm B\kern-.05em{\sc i\kern-.025em b}\kern-.08em
    T\kern-.1667em\lower.7ex\hbox{E}\kern-.125emX}}
\begin{document}

\title{Optimizing Active Learning in Vision-Language Models via Parameter-Efficient Uncertainty Calibration\\
}

\author{\IEEEauthorblockN{Athmanarayanan Lakshmi Narayanan}
\IEEEauthorblockA{\textit{Intel Labs} \\
\textit{Intel Corporation}\\
Santa Clara, USA \\
athma.lakshmi.narayanan@intel.com}
\and
\IEEEauthorblockN{Amrutha Machireddy}
\IEEEauthorblockA{\textit{Intel Labs} \\
\textit{Intel Corporation}\\
Bangalore, IN \\
amrutha.machireddy@intel.com}
\and
\IEEEauthorblockN{Ranganath Krishnan }
\IEEEauthorblockA{\textit{Intel Labs} \\
\textit{Intel Corporation}\\
Hillsboro, USA \\
ranganath.krishnan@intel.com}
}

\maketitle

\begin{abstract}
Active Learning (AL) has emerged as a powerful approach for minimizing labeling costs by selectively sampling the most informative data for neural network model development. Effective AL for large-scale vision-language models necessitates addressing challenges in uncertainty estimation and efficient sampling given the vast number of parameters involved. In this work, we introduce a novel parameter-efficient learning methodology that incorporates uncertainty calibration loss within the AL framework. We propose a differentiable loss function that promotes uncertainty calibration for effectively selecting fewer and most informative data samples for fine-tuning. Through extensive experiments across several datasets and vision backbones, we demonstrate that our solution can match and exceed the performance of complex feature-based sampling techniques while being computationally very efficient. Additionally, we investigate the efficacy of Prompt learning versus Low-rank adaptation (LoRA) in sample selection, providing a detailed comparative analysis of these methods in the context of efficient AL\footnote{Code available at \url{https://github.com/IntelLabs/C_PEAL}}.
\end{abstract}


\section{Introduction}
The ability to efficiently adapt vision-language foundation models \cite{song2023meta,nagar2024zero} to specialized domains efficiently is becoming increasingly important in various real-world applications \cite{hu2023look,lu2024deepseek}. Transfer learning is a crucial technique to leverage these foundation models to domain specific applications in a cost effective manner. Active learning (AL)\cite{settles2009active,ren2021survey} helps in this task by intelligently selecting the most informative data samples for labeling. Active learning can drastically reduce the volume of labeled data required to fine-tune the model, making the adaptation process not only faster but also more cost-effective. AL techniques typically utilize uncertainty sampling or diversity sampling for selecting the data samples for annotation and training the model. However, Vision-Language Models (VLMs) \cite{radford2021learning,ramesh2021zero,singh2022flava} pre-trained on large corpus datasets often produce unreliable probability distributions, leading to poorly calibrated uncertainty estimates \cite{groot-valdenegro-toro-2024-overconfidence, deng2024seeing,liu2024survey,rawte2023survey}. In an AL setup, this can result in repeated annotations of samples from already-known data distribution and overlook truly informative data samples, ultimately wasting the annotation budget. By failing to prioritize diverse and uncertain samples, especially in out-of-distribution (OOD) datasets, the model misses opportunities to maximize the value of each annotation, reducing the overall efficiency of the AL process.

% 2. Current miscalibration in current foundation models? Not sure need to search for some data proving that models are not well calibrated..
%3.Rather than posthoc that uses additional hold out dataset and compute while doing inside AL why this is better
Recent works \cite{xiongcan,fadeeva2023lm, krishnan2024enhancing,groot2024overconfidence} demonstrate the tendency of large language models (LLMs) and VLMs toward overconfidence in their predictions, underscoring the need for effective uncertainty calibration to ensure that model confidence always aligns with accuracy. Moreover, in an AL setup, the scarcity of high-quality annotated datasets, due to the high costs of labeling, makes traditional post-process calibration methods (which require holdout sets), challenging to apply. This situation calls for uncertainty calibration approaches that can function effectively with minimal labeled samples.\\
The key contributions in this paper are as follows:
\begin{itemize}
    %\item We propose a Calibrated Parameter-Efficient Active Learning strategy (C-PEAL) tailored for in vision-language models, an area that remains critically under explored in current research. This methodology is implemented on a popular vision-language model, CLIP \cite{radford2021learning}, for a few-shot classification task, demonstrating its effectiveness in scenarios with limited labeled data where annotation costs are high (scarcity of domain experts, labeling tools, etc). This approach can be extended to various foundation models and datasets.
    \item We introduce a Calibrated Parameter-Efficient Active Learning strategy (C-PEAL) for vision-language models, by proposing a novel trainable uncertainty calibration loss to effectively select informative data samples for few-shot fine-tuning in active learning cycles.
    \item We conduct a thorough evaluation on the effect of choice of fine-tuning strategy (Prompt Learning vs. Low Rank Adaptation (LoRA)) in an active learning setup. To the best of our knowledge, this is the first and essential benchmark, providing insights into the selection of fine tuning strategy and open or closed-source models, with notable performance gains observed using LoRA.
    \item We benchmark the proposed method using four different datasets and three vision backbones, showcasing performance gains across all setups and demonstrating competitive results compared to complex feature-based selection strategies.
\end{itemize}

\begin{figure*}[h]
    \centering
    \includegraphics[width=0.75\linewidth]{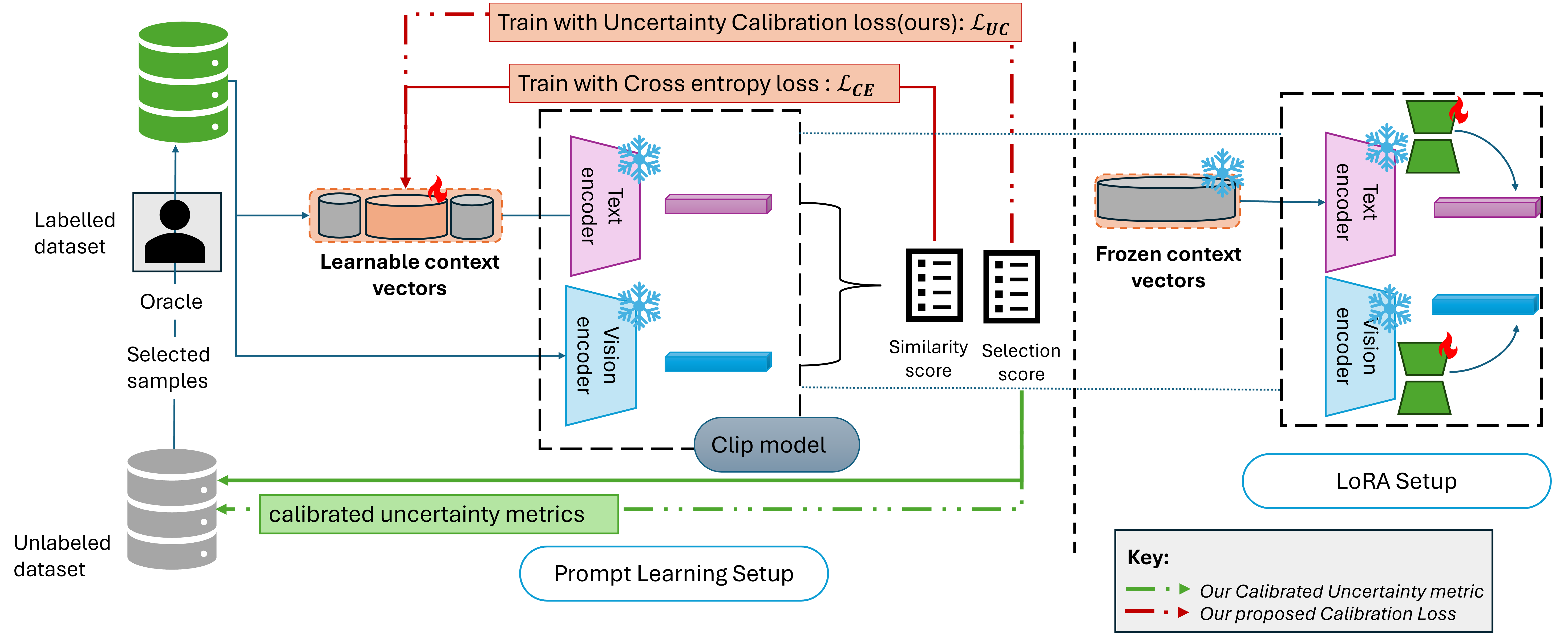} % Placeholder for your figure
     % \captionsetup{font=normal}
    \caption{Overview of Active Learning with Prompt Learning and LoRA setups. In Prompt Learning, only the learnable context embeddings are trained, while in LoRA, only adapter layers for the vision and text encoders are trained. Highlighted in dotted lines is our proposed solution that uses Uncertainty Calibration loss (in red) in addition to Cross Entropy loss to provide better quality sampling scores (in green). }
    \label{fig:mainfigure}
\end{figure*}

\section{Related Work}

Active learning \cite{Xu_2024_CVPR, xie2023active} is an area of research focusing on improving data efficiency and minimizing data annotation cost while maximizing the model performance. The data sampling strategy is broadly classified into diversity-based \cite{cho2024querying,sener2017active} and uncertainty-based \cite{yoo2019learning,ranganathan2017deep} techniques. In diversity-based sampling \cite{cho2024querying}, a subset of samples is chosen such that they span the entire feature space either through clustering or coreset selection techniques. In uncertainty-based sampling, the samples are chosen based on predicted uncertainty metrics such as entropy \cite{wu2022entropy}, mixture density networks \cite{choi2021active} and margin \cite{roth2006margin} to rank the samples. 

Large-scale foundational models pre-trained on a huge corpus of data have enabled state-of-the-art results in various downstream tasks. With the abundance of unlabeled data available, selecting high-quality data for fine-tuning foundational models to specific domains is a critical task \cite{wan2023survey, gupte2024revisiting, longpre2022active,narayanan2024parameterefficientactivelearningfoundational,machireddy2024source}. Approaches such as Prompt learning  coupled with class-balancing\cite{bang2024active,zhou2022learning} and DropQuery \cite{gupte2024revisiting} have been proposed for improving the sample selection for foundation models. Foundational models perform well in the one-shot \cite{jin2022one} and few-shot learning \cite{Han_2024_CVPR, Silva-Rodriguez_2024_CVPR} scenarios. However, identifying and annotating these fewer and most useful data samples is crucial, which we address with our C-PEAL approach. Further, fine-tuning all the model parameters is compute intensive and time consuming. To overcome this, parameter-efficient fine-tuning techniques \cite{Peng_2024_CVPR, Guo_2024_CVPR} help utilize the knowledge from the foundation model by introducing a small set of trainable parameters which can be fine-tuned for specific tasks. Extensions of prompt learning, such as CoOp \cite{zhou2022learning}, Maple \cite{khattak2023maple}, and CoCoOp \cite{zhou2022conditional} have also been introduced. However, these methods mainly focus on few-shot adaptation and assume the availability of annotated data samples. In this work, we address selecting the most useful data samples for the few-shot adaptation of foundation models that maximize model performance as compared to randomly selected data.
%generalization to unseen classes, without addressing AL scenarios. Other extensions to prompt learning, such as \cite{khattak2023maple, khattak2023self}, have been introduced for multimodal prompt learning and continual learning. While we focus on prompt learning in this paper, these extensions could be applied ad hoc in conjunction with our approach.

%uncertanity based methods and techniques.
%1. discuss Avuc paper here and show how it is different.
%. Pick other uncertainity related calibration methods.

Calibration training approaches have proven effective in enabling models to provide high-quality uncertainty estimates, thereby increasing their reliability. The accuracy versus uncertainty calibration (AvUC) method introduced in \cite{krishnan2020improving} leverages the relationship between accuracy and uncertainty to enhance model calibration, particularly for Bayesian neural networks. An extension of this work, Soft-AvUC \cite{cain2021soft}, adapts the approach for standard deep neural networks by utilizing soft binning and gradient-driven optimization for well-calibrated predictions. However, these techniques rely on threshold settings to identify and distinguish levels of uncertainty, which can be challenging, especially in active learning settings where determining thresholds for each AL cycle will become complex. We address this challenge by proposing a novel threshold-free uncertainty calibration loss. Our approach relies on first principles, encouraging the model to provide reliable uncertainty estimates along with the predictions by performing min-max optimization of predictive probability and uncertainty scores based on correct and incorrect predictions during training. This method eliminates the need for threshold settings, simplifying the calibration training and enhancing its effectiveness in active learning scenarios.

%which has not yet been thoroughly explored in an AL setting, where determining thresholds for each AL cycle can be challenging. In our work, we address this by avoiding threshold-based uncertainty optimization altogether. Instead, we leverage a threshold-free learning-driven calibration mechanism that dynamically adjusts uncertainty estimates and associated weights. This approach has not been previously explored within an AL context, presenting a novel application of uncertainty calibration. 

\section{Problem Formulation}

\subsection{Parameter Efficient Active Learning}
In AL, the model iteratively selects the most informative samples from an unlabeled dataset \( \mathcal{D_U} \) to maximize performance and minimize labeling costs. At each cycle \( t \), a subset \( \mathcal{B}_t \subset \mathcal{D_U} \) is selected and added to the labeled pool \( \mathcal{D_L} \) by a human oracle. This process continues in a batch fashion until the budget is exhausted or the model converges.

Active learning has also been investigated for vision-language models like CLIP \cite{radford2021learning} within a parameter-efficient fine-tuning framework. Due to the impressive zero- and few-shot performance of foundation models, samples are selected in a few-shot manner, with each active learning cycle selecting \( B_{t} = K \), where \( K \) is the number of classes. To address the high training costs associated with the large number of trainable parameters in foundation models, parameter-efficient fine-tuning (PEFT) methods, such as Prompt learning \cite{zhou2022learning,lester2021power} or Low-rank adaptation (LoRA) \cite{hu2022lora}, can be used to train a small subset of parameters. The overall active learning process under both the prompt learning and LoRA setups is illustrated in Figure \ref{fig:mainfigure}.

% 1. Find some papers that talk about lora and other isn active learning setup
% 2. breifly explain. 
% 3. Now talk about how we need to use impressive performance of few shot founation mdoels.
% 4. Hence it is imperive this study is used irrespective of choice of peft method.
% 5. why is AL still relavent for foundation?
% 6. In this setup we do classification..explain.
% 7.Also talk about AL in VLMS and LLLMS. primartily prevoius on seperate tasks object detection..etc..
% 8.This is a first step toward understndin VLMS and future downstream taks.
% use this:
% Vision-Language Models (VLMs) like CLIP have been instrumental in performing few-shot learning tasks. CLIP, which combines visual and text representations, enables models to understand images and associate them with textual descriptions. This allows for effective zero-shot and few-shot classification, where models trained on vast amounts of data can generalize to unseen classes with minimal examples.
% In few-shot setups, such models are fine-tuned using only a few labeled examples per class. The power of CLIP lies in its ability to use textual prompts to describe classes and match those descriptions with image embeddings. As a result, few-shot learning tasks can be efficiently handled without the need for extensive labeled data.
\textbf{Prompt Learning} Prompt learning operates by designing templates that transform input data into textual prompts interpretable by the model. A standard approach involves using natural language templates of the form: ``A photo of a \( K_{\text{name}} \)" where \( K_{\text{name}} \) represents the label name associated with input \( x \). These prompts serve as guiding structures for the model’s inference process, enhancing its generalization capability, particularly when labeled data is scarce. Within a prompt, specific tokens are designated as trainable, while others remain fixed. Notably, class tokens, as well as components such as the vision encoder and text encoder, remain frozen. The trainable tokens are transformed into embeddings, denoted as \( V_{\text{PE}} \), which are subsequently injected into the embedding vectors at intermediate layers. 

Following the approach in \cite{bang2024active}, our setup also employs trainable context vectors, denoted as \( \text{ctx} \), with a fixed size \( \text{ctx} = 16 \). Unlike prior work that utilizes description augmentation (i.e., employing multiple template variations per class), we do not incorporate such augmentations, as it is beyond the scope of this work. Consequently, the trainable parameters in our method take the form of a tensor with dimensions \( K \times 16 \times E \), where \( K \) represents the number of classes and \( E \) denotes the embedding dimension.

\textbf{Low-Rank Adaptation (LoRA)} LoRA keeps the pretrained model weights unchanged and adds trainable matrices that decompose the ranks within each transformer layer. More specifically, instead of directly fine-tuning \( W \), which is the weight matrix of learnable parameters in the CLIP model with dimensions \( W \in \mathbb{R}^{d \times k} \), a low-rank decomposition is introduced: \( \Delta W = A B \), where \( A \in \mathbb{R}^{d \times r} \), \( B \in \mathbb{R}^{r \times k} \), and \( r \ll \min(d, k) \). Thus, the adapted weight matrix is updated as \( W' = W + A B \). This approach significantly reduces the number of parameters that need to be trained for downstream tasks, providing a more efficient alternative to training all model weights. We inject LoRA adapters into the \( q \), \( k \), and \( v \) attention layers within each transformer block of both the vision and text encoders, while keeping the prompt embeddings from the template and the base model frozen. Unlike in \textit{Prompt Learning}, the learnable parameter size in LoRA is unaffected by \( K \). By manipulating rank of the LoRA adapters (r=2 or r=4) we find that the number of trainable parameters are in some cases lesser than \textit{Prompt Learning.}

\subsection{Optimizing Active Learning Complexity}
Active learning strategies based on feature distance metrics \cite{ash2019deep, sener2017active, krishnan2021robust} often outperform strategies using single-shot uncertainty metrics. However, these methods are computationally costly, as calculating pairwise distances over large datasets becomes cumbersome with extensive unlabeled samples. The runtime complexity of feature embedding-based AL increases substantially with the size of the unlabeled pool in BADGE\cite{ash2019deep}, and with the sizes of both the labeled and unlabeled pools in Coreset\cite{sener2017active}. Table \ref{table:relative_time} shows the runtime for the selection strategies across datasets during the first AL cycle, with the largest number of unlabeled data points. Experiments were conducted on a system with an Intel Xeon Gold CPU (128 cores) and one A100 80GB GPU. Achieving parameter- and runtime-efficient sampling requires considering both training parameters and sampling complexity. Hence, lightweight uncertainty-based methods, such as Entropy, play a significant role, reinforcing our focus on calibrating uncertainty metrics as a well-justified choice. As explained in Sec.\ref{sec:experiments}, our method demonstrates performance that is either superior in most cases or competitively comparable with BADGE while being compute-efficient. In contrast, Coreset performs worse than random sampling in high dimension feature embedding space.

\begin{table}[h]
\centering
\begin{tabular}{lccc}
\toprule
Dataset & C-PEAL (Ours)/Entropy/Softmax & Coreset & BADGE \\
\midrule
Caltech101 & 1.00x & 1.23x & 107.42x \\
Eurosat & 1.00x & 1.36x & 54.54x \\
DTD & 1.00x & 1.71x & 24.24x \\
Oxford Pets & 1.00x & 1.55x & 64.41x \\
\bottomrule
\end{tabular}
\caption{Relative time taken by each selection method for different datasets in the first AL cycle using the ViT-B/32 backbone. C-PEAL (Ours) is set as the baseline (1x) and has a runtime comparable to Entropy or Softmax-based sample selection. The table compares Coreset and BADGE with our method, highlighting the quadratic growth in BADGE’s selection time as the number of unlabeled samples increases.}
\label{table:relative_time}
\end{table}

% \begin{figure}[ht]
%     \centering
%     \includegraphics[width=0.4\textwidth]{sec/pics/runtimeplots.png} % Adjust the width as needed
%      \captionsetup{font=small}
%     \caption{Time taken by each selection method for different datasets in the first AL cycle using the ViT-B/32 backbone. The values indicate the time (in seconds) required for the C-Peal(Ours), Coreset, and BADGE selection methods across the datasets. The plot highlights the quadratic growth of the BADGE selection time as the number of unlabeled samples increases}
%     \label{fig:time_heatmap}
% \end{figure}

\section{Methodology}
We propose parameter-efficient active learning with calibrated uncertainty estimates for effectively selecting fewer data samples to fine-tune foundation models. We introduce a novel uncertainty calibration loss to encourage the model to learn to provide well-calibrated uncertainty estimates during the AL iterations. Unlike previous post-processing calibration methods \cite{guo2017calibration, naeini2015obtaining} that utilize a hold-out test set to calibrate the model after training, our approach integrates calibration directly within the active learning process, thereby reducing dependency on additional external quality data. Our proposed method aims to calibrate the sample selection function directly by leveraging the already labeled train set in each AL Cycle. This approach allows the model to make informed sample selections based on calibrated uncertainty, ultimately improving the relevance and quality of predictions on unseen data.

We illustrate the proposed approach on two distinct PEFT methods: Prompt learning \cite{lester2021power} and Low-rank adaptation \cite{hu2022lora}. These fine-tuning methods enable efficient adaptation of foundation models with minimal parameter updates, preserving the overall structure and robustness of the pretrained model. Figure \ref{fig:mainfigure} shows the configuration for the CLIP model in the parameter-efficient AL setup.

%To better understand the approach, and as shown in Figure \ref{fig:al_prompt_learning}, we set up a configuration for the CLIP model that incorporates two distinct parameter-efficient fine-tuning methods: prompt learning and low-rank adaptation. These methods enable efficient adaptation of the model with minimal parameter updates, preserving the overall structure and robustness of the pretrained model.

%Our proposed method focuses on calibrating uncertainty and, unlike previous approaches, leverages the learned parameters during training to optimize the sampling selection process for inference on an unlabeled dataset. This novel approach allows the model to make informed sample selections based on calibrated uncertainty, ultimately improving the relevance and quality of predictions on unseen data.

\subsection{Calibrated Uncertainty}
Let \( B \) denote the mini-batch of predictions. Each instance in the mini-batch is associated with an index \( i \), where \( i = 1, 2, \dots, |B| \). Let \( y_i \) denote the ground-truth label for the \( i \)-th instance in \( B \), and let \( \hat{y}_i \) denote the model's predicted label for the \( i \)-th instance in \( B \). We define two subsets of indices in \( B \) as follows:
\[
\mathcal{I} = \{i \in \{1, 2, \dots, |B|\} \mid \hat{y}_i \neq y_i \}
\]
\[
\mathcal{C} = \{j \in \{1, 2, \dots, |B|\} \mid \hat{y}_j = y_j \}
\]
\( \mathcal{I} \) is the set of indices corresponding to incorrect predictions, where the predicted label \( \hat{y}_i \) does not match the ground-truth label \( y_i \). \( \mathcal{C} \) is the set of indices corresponding to correct predictions, where the predicted label \( \hat{y}_j \) matches the ground-truth label \( y_j \).

Uncertainty-based sample selection techniques rely on reliable metrics to prioritize samples with higher uncertainty estimates for labeling, thereby enhancing the model's learning process. For a well-calibrated model, it is crucial to indicate higher uncertainty for unseen data samples, indicating areas of model unknowns. Conversely, the model should indicate low uncertainty for data samples from the previous seen data distribution, where it tends to make accurate predictions. This calibration of uncertainty estimation ensures that the active learning process is efficient and effective, focusing labeling efforts on the most informative samples. There are various uncertainty metrics \cite{shannon1948entropy,houlsby2011bayesian,mukhoti2023deep} that have been used for sample selection in active learning. To select informative samples, we leverage predictive entropy (Equation~\ref{eqn:entropy}) as a robust and computationally efficient uncertainty measure. However, our proposed loss is versatile and can be applied to calibrate any uncertainty function.

%\textcolor{red}{Needs more points on calibration and some explanations based on equations.}

%The entropy of the prediction is given by,
\begin{align}
\label{eqn:entropy}
\text{H}(p_i) &= -\sum_{k=1}^{K} p_i^{(k)} \log(p_i^{(k)} )
\end{align}
where $K$ is the number of classes, $p_i^{(k)}$ is the probability of class $k$ for sample $i$.
Then we can define $\text{unc}_{\text{incorrect}}$ and $\text{unc}_{\text{correct}}$ as follows,
\begin{align}
\text{unc}_{\text{incorrect}} &= \left\{ \text{H}(p_i) \mid i \in \mathcal{I} \right\} \\
\text{unc}_{\text{correct}} &= \left\{ \text{H}(p_i) \mid i \in \mathcal{C} \right\}
\end{align}

We expect to be confident on accurate predictions and obtain likely higher uncertainty on inaccurate predictions. Therefore we define incorrect loss $\mathcal{L}_I$ and correct loss $\mathcal{L}_C$ as follows,

\begin{align}
\mathcal{L}_{\text{I}} &= \frac{1}{|\text{unc}_{\text{incorrect}}|}
\sum_{i=1}^{|\text{unc}_{\text{incorrect}}|} -\log(\tanh(\text{unc}_{\text{incorrect}}^{(i)}) + \epsilon) \\
\mathcal{L}_{\text{C}} &= \frac{1}{|\text{unc}_{\text{correct}}|}
\sum_{j=1}^{|\text{unc}_{\text{correct}}|} -\log(1 - \tanh(\text{unc}_{\text{correct}}^{(j)}) + \epsilon)
\end{align}

The loss functions \( \mathcal{L}_{\text{I}} \) and \( \mathcal{L}_{\text{C}} \) are designed to penalize the model based on its uncertainty values for incorrect and correct predictions, respectively. The constant $\epsilon$ ($10^{-6}$) is added to prevent numerical instability by avoiding $\log(0)$, ensuring stable gradients, and mitigating floating-point precision issues.
The hyperbolic tangent function \( \tanh(u_i) \) is applied to the uncertainty values, scaling them to the range \([0, 1]\). This makes the uncertainty values easier to interpret within a bounded range for min-max optimization. 

For incorrect predictions, the model is encouraged to assign high uncertainty values $(\text{unc} \to 1)$. This penalizes cases where uncertainty is low, pushing \( \tanh \) component higher. This helps the model indicate a high level of uncertainty when it is likely wrong. For correct predictions, the model should ideally be confident, so it is penalized if it has high uncertainty. This is achieved by minimizing \( \tanh(\text{unc}_{\text{correct}})\), encouraging the uncertainty values for correct predictions to be $(\text{unc} \to 0)$. This loss setup encourages well-calibrated confidence by aligning the model’s certainty with its actual performance.

\begin{figure}[h]
    \centering
\includegraphics[width=0.9\columnwidth]{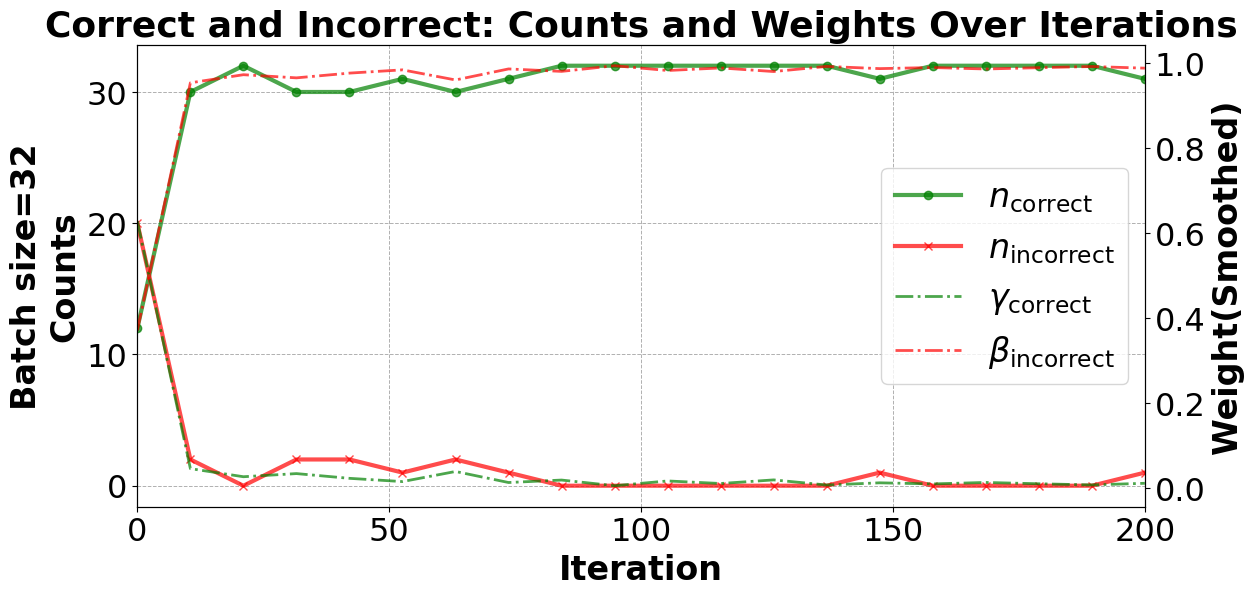} % Replace with the path to your image
 % \captionsetup{font=small}
\caption{The number of correct (\(n_{\text{correct}}\)) and incorrect (\(n_{\text{incorrect}}\)) predictions over the training iterations during AL cycle 1. As training progresses, the incorrect counts decrease quickly after a few iterations, requiring the associated loss weight (\(\beta_{\text{incorrect}}\)) to increase proportionally. Conversely, the weight for correct predictions (\(\gamma_{\text{correct}}\)) should start high and decrease as the number of correct predictions increases. This demonstrates how the loss weights adapt dynamically to balance the contributions from correct and incorrect predictions.}
    \label{fig:correctnincorrect}
\end{figure}

\subsection{Uncertainty Calibration loss}
\label{sec:uncertainty calibration loss}
We train the model with a secondary loss for calibration. During the initial training epochs, the model generates a higher number of incorrect samples, but this number decreases over time as the model learns and produces more correct samples. This trend is more pronounced in this case due to the inherent zero-shot capabilities of foundation models, which quickly improve with training, as shown in Figure \ref{fig:correctnincorrect}. This imbalance causes the secondary losses to have high variance and increases the time taken for model convergence. Hence, with this observation, we propose a weighting scheme within each mini-batch to regularize training by balancing the losses for correct and incorrect samples.
\begin{align}
\mathcal{L}_{\text{calib}} &= \gamma_{correct} \cdot \mathcal{L}_{\text{C}} + \beta_{incorrect} \cdot \mathcal{L}_{\text{I}} \\
\gamma_{correct} &= \frac{n_{\text{incorrect}}}{n_{\text{correct}} + n_{\text{incorrect}}} \\
\beta_{incorrect} &= \frac{n_{\text{correct}}}{n_{\text{correct}} + n_{\text{incorrect}}}
\end{align}

Loss function can be optimized using standard gradient descent, allowing the model to learn well-calibrated uncertainty estimates while also enhancing the primary task prediction accuracy ($\mathcal{L}_{CE}$).
\begin{align}
\mathcal{L}=\mathcal{L}_{CE}+\alpha \mathcal{L}_{\text{calib}}
\end{align}
Similar to other multi-task learning approaches, the parameter \( \alpha \) serves as a weighting factor for the calibration loss. It is linearly annealed over training to gradually increase its influence, enabling the model to balance calibration and task accuracy progressively. It is linearly annealed over training to gradually increase its influence, enabling the model to balance calibration and task accuracy progressively. The final value of \( \alpha \) is selected via hyperparameter grid search and primarily affects the speed of convergence, with minimal impact on final accuracy.

Our proposed algorithm utilizes calibrated uncertainty in the AL loop. Unlike other baselines, our method introduces only minimal changes to the learning loss, adding negligible computational overhead to capture statistics that link accuracy with confidence. These learned components aid in sample selection during the inference stage. At each step, the model selects the sample with the highest entropy, weighted by the calibration factor. For completeness, the overall active learning loop in shown in Algorithm \ref{alg:calibrated_active_learning}.

% \begin{algorithm}
% \caption{Calibrated Uncertainty-Based Active Learning}
% \label{alg:calibrated_active_learning}
% \begin{algorithmic}
% \STATE Initialize Seed \( S \)
% \STATE Initialize model \( M \) with learnable parameters \( P \), where:
% \STATE \quad Choose \( P \in \{\theta_{\text{LoRA}}, \theta_{\text{Prompt}}\} \)
% \STATE \quad \textbf{LoRA}: \( \theta_{\text{LoRA}} = \{W + \Delta W \mid \Delta W = A B^\top\} \)
% \STATE \quad \quad \( A \in \mathbb{R}^{d \times r}, \quad B \in \mathbb{R}^{d \times r}, \quad r \ll d \)
% \STATE \quad % \vline
% \STATE \quad \textbf{Prompt Learning}: \( \theta_{\text{Prompt}} = \{V_\text{{Prompt embeddings}}\} \)
% \STATE Initialize labeled set \( D_{L} \) and unlabeled set \( D_{U} \)
% \FOR{each iteration}
%     \STATE Train \( M \) on \( D_{L} \) with loss:
%     \STATE \quad \( \mathcal{L} = \mathcal{L}_{\text{CE}} + \alpha \mathcal{L}_{\text{calib}} \)
%     \FOR{each sample \( x_i \in D_{U} \)}
%         \STATE Compute uncertainty \( U(x_i) \)
%         \STATE Compute weighting factor \( W(x_i) \)
%         \STATE Compute calibrated uncertainty \( U'(x_i) = W(x_i) \cdot U(x_i) \)
%     \ENDFOR
%     \STATE Select sample with highest \( U'(x_i) \)
%     \STATE Label selected sample and add to \( D_{L} \)
% \ENDFOR
% \end{algorithmic}
% \end{algorithm}

\begin{algorithm}
\caption{Calibrated Uncertainty-Based Active Learning}
\label{alg:calibrated_active_learning}
\Repeat{For $\text{Seed}=S1,S2,S3$, $\text{AL Cycle}=8$ }{

    \ForEach{AL Cycle}{
        Initialize seed \( S \)\;
        Initialize model \( M \) with learnable parameters \( P \), where:\;
        \Indp
            Choose \( P \in \{\theta_{\text{LoRA}}, \theta_{\text{Prompt}}\} \)\;
            
            \textbf{LoRA}: \( \theta_{\text{LoRA}} = \{W + \Delta W \mid \Delta W = A B^\top\} \)\;
            % \Indm
            \textbf{Prompt Learning}: \( \theta_{\text{Prompt}} = \{V_\text{PE}\} \)\;
        \Indm
        
    Initialize labeled set \( D_{L} \) and unlabeled set \( D_{U} \)\;
        
            1) Train \( M \) on \( D_{L} \) with calibrated loss:\;
            \Indp
                a) Compute weighting factors \(\gamma\) and \(\beta\) for mini-batch\;
                b) Set \(\alpha\)\;
                c) Compute \( \mathcal{L} = \mathcal{L}_{\text{CE}} + \alpha \mathcal{L}_{\text{calib}} \)\;
            \Indm
            2) Sample Selection:\;
            \ForEach{sample \( x_i \in D_{U} \)}{
                Compute calibrated entropy \( H(x_i) \)\;
            }
            Select $B$ samples with the highest \( H(x_i) \)\;
            
            3) Label selected sample and add to \( D_{L} \)\;
    }
}
\end{algorithm}

\subsection{Uncertainity Calibration in AL}
% Active learning relies heavily on uncertainty metrics to prioritize which samples to label. Several uncertainty based acquisition functions have been used in literature namely, Entropy, least confidence, Margin which are probabilistic. Other probabilistic methods such as BALD maximize the mutual information between the model’s predictions and its parameters. For a comprehensive overview of work in active learning, we refer to \cite{ren2021survey,settles2009active}.

\begin{figure}[ht]
    \centering
    % First Subfigure
    \begin{subfigure}{0.49\columnwidth}
        \centering
        \includegraphics[width=\linewidth,height=0.16\textheight]{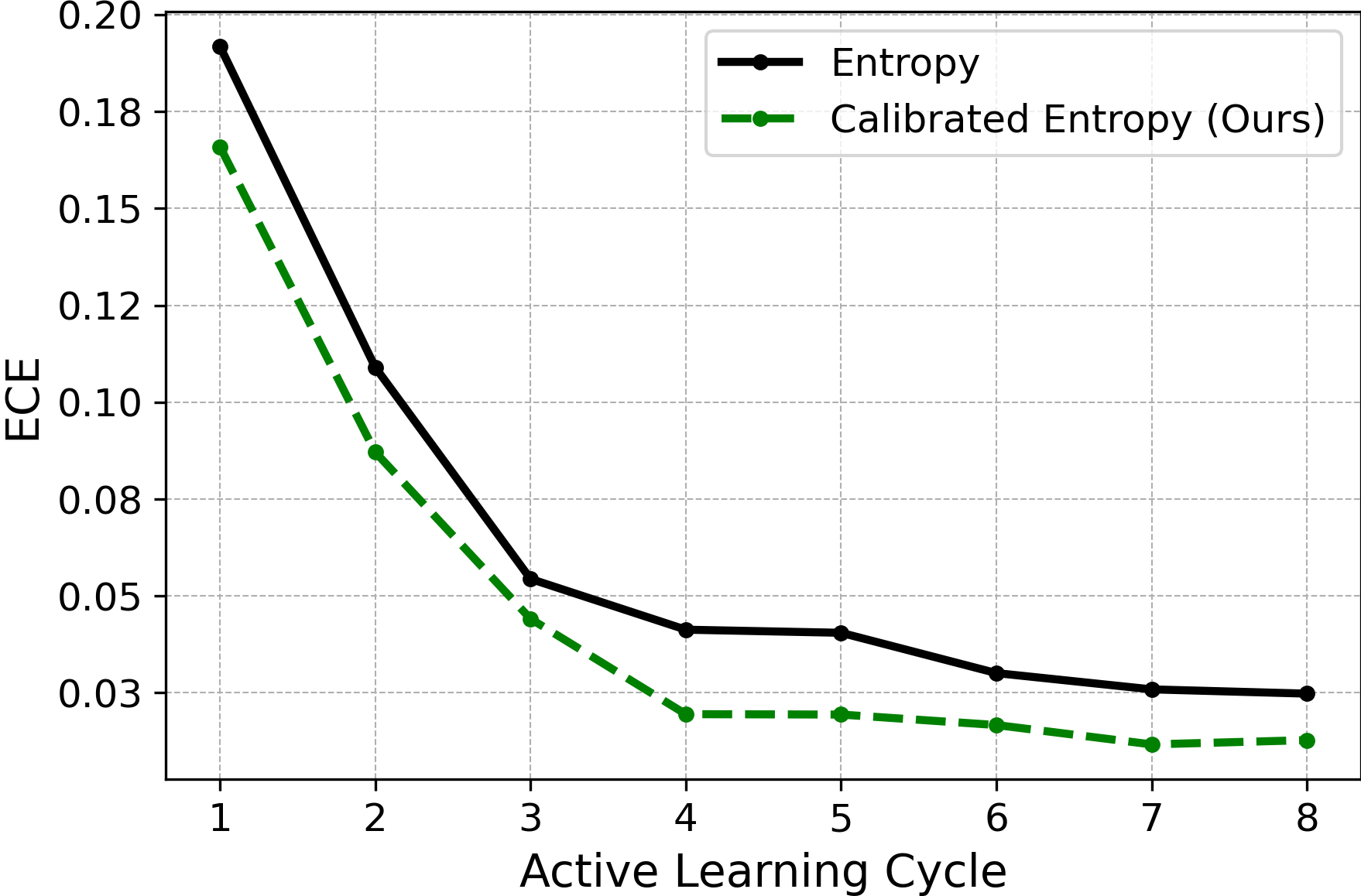}
        \caption{Caltech101}
        \label{fig:plot1}
    \end{subfigure}
    \hfill
    % Second Subfigure
    \begin{subfigure}{0.49\columnwidth}
        \centering
        \includegraphics[width=\linewidth,height=0.16\textheight]{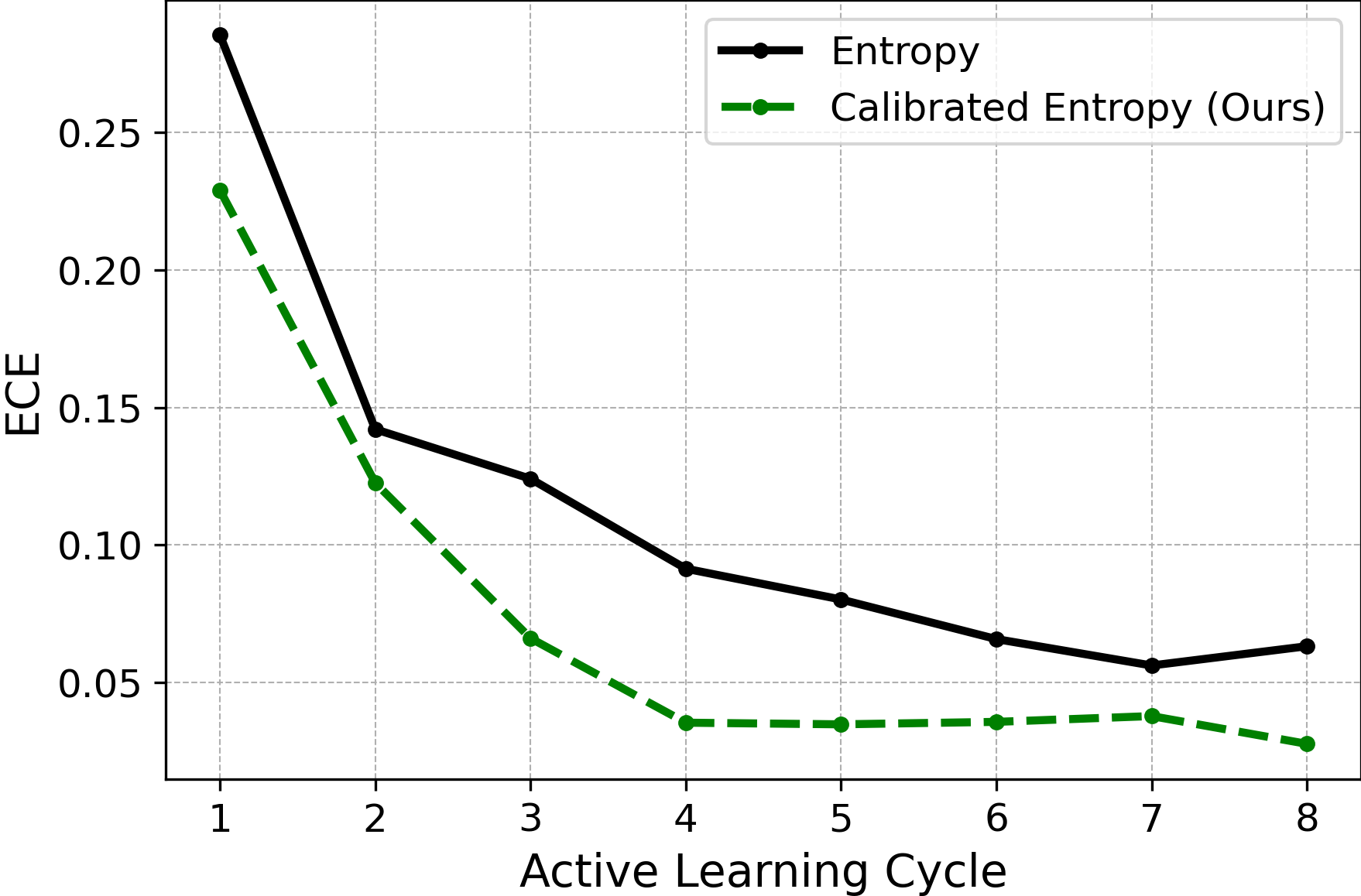}
        \caption{Oxford Pets}
        \label{fig:plot2}
    \end{subfigure}
    \hfill
    \begin{subfigure}{0.49\columnwidth}
        \centering
        \includegraphics[width=\linewidth,height=0.16\textheight]{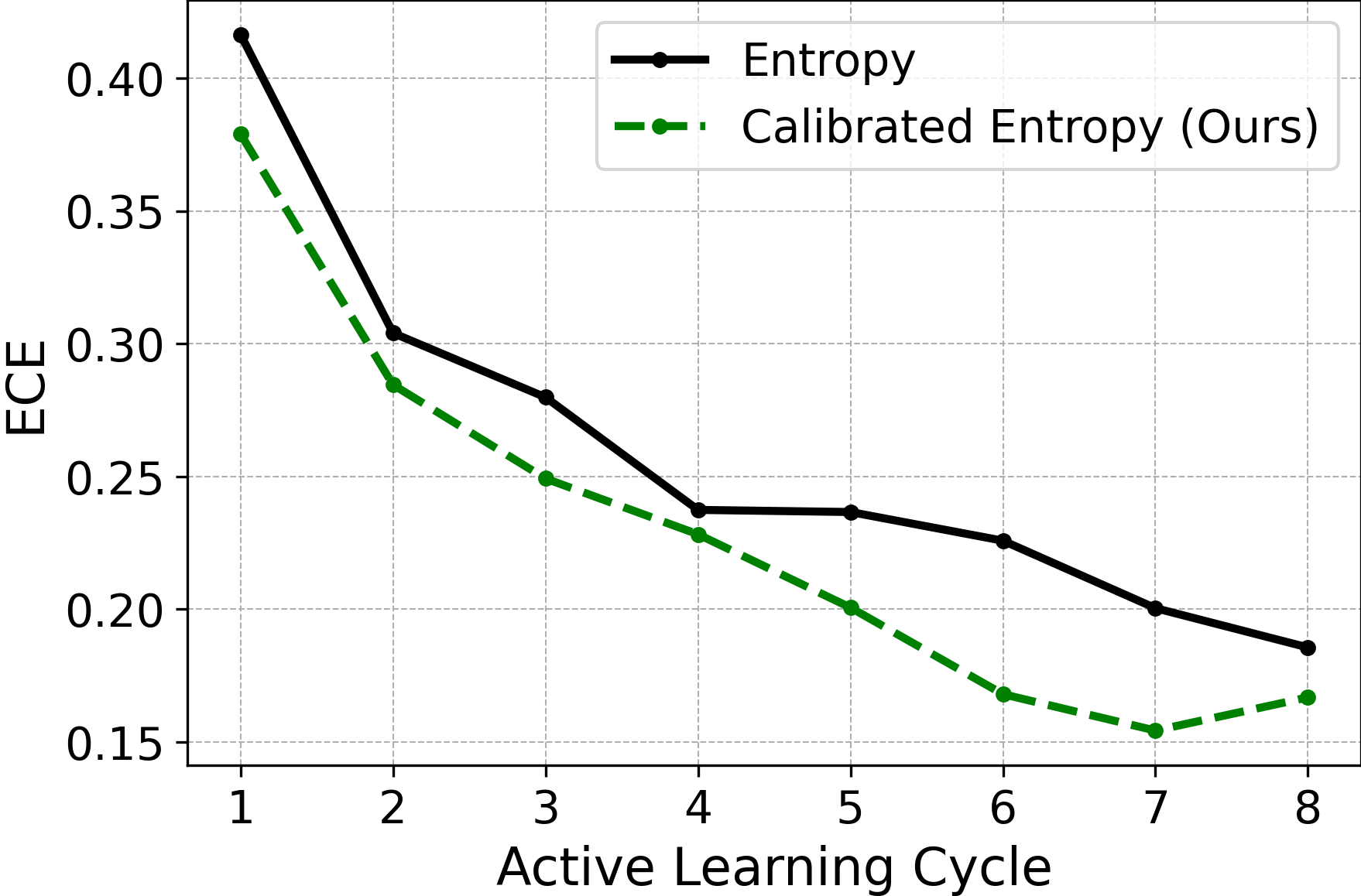}
        \caption{DTD}
        \label{fig:plot2}
    \end{subfigure}
    \hfill
    \begin{subfigure}{0.49\columnwidth}
        \centering
        \includegraphics[width=\linewidth,height=0.16\textheight]{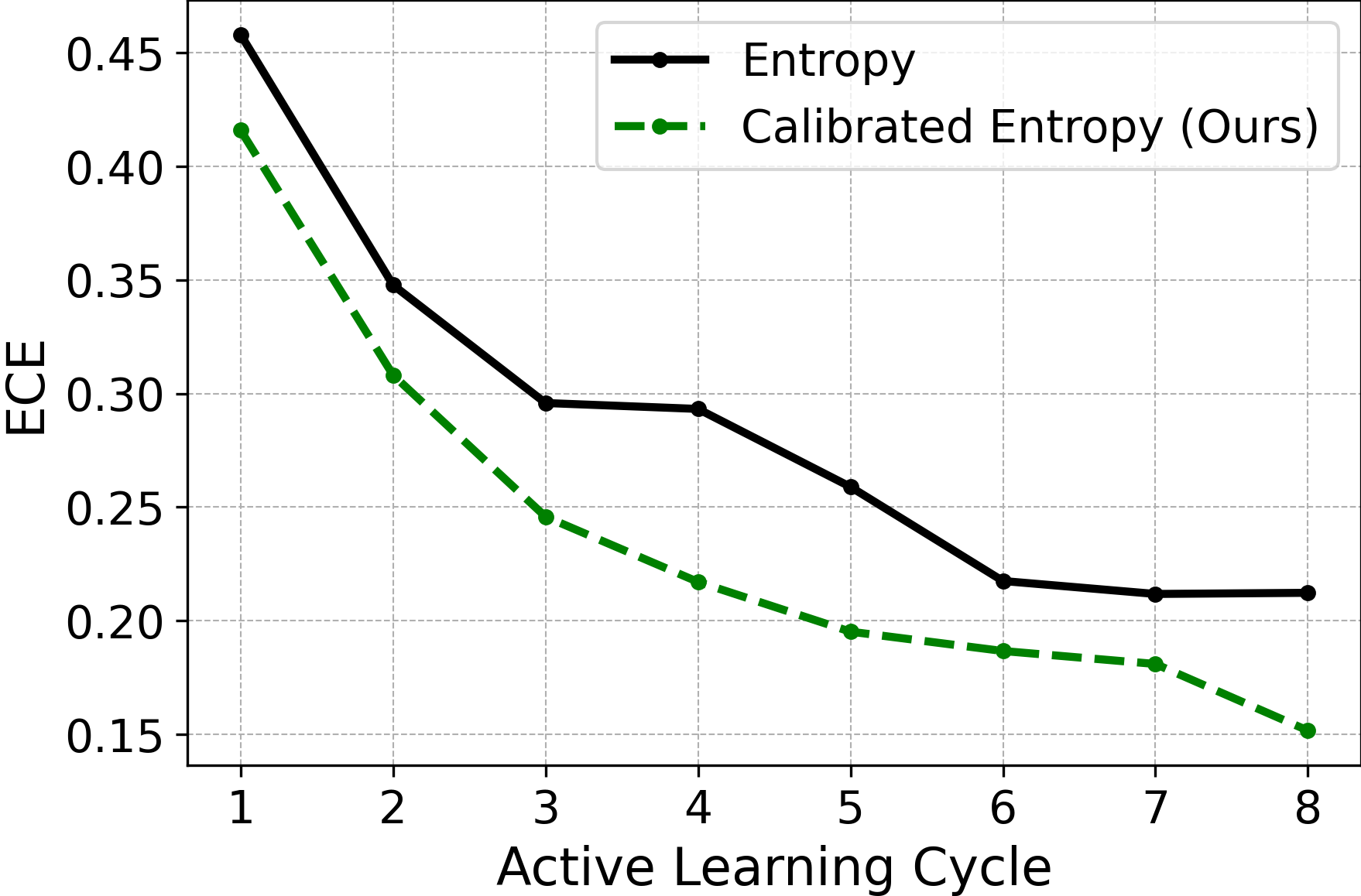}
        \caption{Eurosat}
        \label{fig:plot2}
    \end{subfigure}

     % \captionsetup{font=small}
    \caption{ Expected Calibration Error (ECE)$\downarrow$ as a function of active learning cycles. Lower ECE is desired for better uncertainty calibration in active learning. The plots show results for ViT-B/32 on (a) Caltech101 and (b) Oxford Pets (c) DTD (d) EuroSAT .}
    \label{fig:eceplots}
\end{figure}

Expected Calibration Error (ECE) \cite{naeini2015obtaining} measures the difference between predicted probabilities and actual outcomes. In a well-calibrated model, ECE decreases over training, as shown in Figure \ref{fig:eceplots}, indicating alignment between confidence scores and true performance. Lower ECE indicates better sample quality, as informative samples speed up convergence to optimal accuracy on the full data.

Our proposed method significantly reduces ECE in each active learning cycle, consistently achieving lower ECE values compared to baseline entropy-based selection as shown in Figure \ref{fig:eceplots}. This validates and highlights the effectiveness of the uncertainty calibration loss in AL, demonstrating substantial benefits in model calibration.

\section{Experiments and discussion}
\label{sec:experiments}
\begin{table*}[h!]
\centering
\small
\renewcommand{\arraystretch}{0.8} % Adjust the row height (less than 1 reduces the height)
\begin{tabular}{@{}lcccccc@{}}
\toprule
\textbf{Method} & \textbf{DTD} & \textbf{Oxford Pets} & \textbf{EuroSAT} & \textbf{Caltech101} & \textbf{Avg Acc (↑)} \\ 
\midrule
CLIP (zero-shot) & 44.5 & 87.0 & 49.4 & 87.9 & 67.2 \\
Random & 58.77 ± 1.94 & 78.30 ± 0.74 & 77.62 ± 1.12 & 89.55 ± 1.00 & 76.06 \\ 
\midrule
\multicolumn{6}{l}{\textit{$O(n)$ Methods}} \\
\cmidrule(lr){1-6}
Entropy \cite{bang2024active} & 59.73 ± 1.96 & 80.44 ± 1.24 & 80.80 ± 2.88 & 92.41 ± 0.50 & 78.34 \\
Softmax \cite{ren2021survey} & 58.87 ± 0.27 & 80.70 ± 0.44 & 79.11 ± 0.25 & 92.14 ± 0.12 & 77.70 \\
Margin \cite{ren2021survey} & 60.22 ± 0.38 & 80.83 ± 1.74 & 79.01 ± 0.01 & 92.54 ± 0.37 & 78.15 \\
\midrule
\multicolumn{6}{l}{\textit{$O(n^2)$ Methods}} \\
\cmidrule(lr){1-6}
Coreset \cite{bang2024active,sener2018active} & 55.77 ± 1.33 & 76.84 ± 1.10 & 77.50 ± 4.64 & 89.96 ± 0.03 & 75.01 \\ 
BADGE \cite{bang2024active,ash2019deep} & \underline{60.28 ± 0.75} & 80.22 ± 1.69 & 81.98 ± 0.81 & 92.21 ± 0.92 & 78.67 \\ 
\midrule
\multicolumn{6}{l}{\textit{Our Method} $O(n)$} \\
\cmidrule(lr){1-6}
C-PEAL  & \textbf{61.0 ± 0.09} & 81.34 ± 1.00 & 82.75 ± 0.31 & \underline{92.64 ± 0.35} & 79.43 \\ 
+ ANN  & 60.17 ± 1.59 &  \underline{82.10 ± 0.69} & \underline{83.69 ± 0.69} & 92.49 ± 0.33 &  \underline{79.61} \\  
+ INTERW & 60.05 ± 1.27 &\textbf{ 82.27 ± 0.17} & \textbf{84.01 ± 0.12} & \textbf{94.10 ± 0.28} & \textbf{80.10} \\ 

\midrule
Full data & 74.7 & 89.3 & 94.5 & 94.4 & 88.125 \\
\midrule
\% of samples & 13.00 & 10.05 & 0.590 & 19.00 & - \\
\midrule
$+\Delta$ Entropy & 1.27 & 1.83 & 3.21 & 1.69 & 2.00 \\
\bottomrule

\end{tabular}

% \captionsetup{font=normal}
\caption{Final accuracy on four downstream tasks with ViT-B/32 image encoder. Final Accuracy is the accuracy after eight rounds, using a prompt learning setup and Avg Acc is the average of final accuracies of four datasets. +INTERW refers to weight balancing as described in Section. \ref{sec:uncertainty calibration loss}. Best results are highlighted in \textbf{bold}, and the second-best results are \underline{underlined}.
}
\label{table:prompt_learning_vit32_main}
\end{table*}
% Write two paras on expiremental setup and detilas on training.
We perform empirical evaluation on diverse datasets and model architectures. We select four major datasets: EuroSAT \cite{helber2019eurosat}, Caltech101 \cite{fei2006one}, DTD \cite{cimpoi14describing}, and Oxford Pets \cite{parkhi12aoxfpets}, each with different class imbalance ratios. For model generalization, we employ three distinct vision backbones in the CLIP model \cite{radford2021learning}: ResNet-50 \cite{he2016deep}, ViT-B/16 \cite{dosovitskiy2021an} (Small), and ViT-B/32 \cite{dosovitskiy2021an} (Large). We train the CLIP model using stochastic gradient descent (SGD) for both prompt tuning and LoRA setup. Following the experimental setup of \cite{bang2024active}, the model is trained with a learning rate of 0.002, for 200 epochs following a cosine annealing schedule over a maximum of 5 epochs and weight decay of 0.0005. An initial warm-up phase with a constant learning rate of $1 \times 10^{-5}$ is applied for the first epoch.

For the prompt learning setup, we maintain the context vectors similar to \cite{bang2024active}, with the context set to $\text{ctx} = 16$ vectors. In addition, we apply a calibration loss by conducting a grid search for $\alpha$ values in the range of 0.1 to 1.0, selecting the best-performing value. For the LoRA setup, we utilize a rank of $r = 2$ or $4$ to reduce the parameter count. This choice helps to balance model complexity and efficiency during training. 
% More details on individual models and parameters are provided in the Appendix Sec. \ref{appendixsec:Experimental setup}.

\subsection{Active learning setup and baselines}
We conduct experiments over eight AL cycles for each model and each dataset. To ensure reproducibility, we perform 3 runs with different random seeds, keeping the same seeds consistent across the experimental setup. All reported results are the mean values over the three runs, accompanied by the standard deviation.

In each AL cycle, the model is reinitialized with random weights to introduce variability and maintain consistency with the experimental setup used in existing literature \cite{bang2024active}. Given the single/few-shot learning setup, only $B = K$ samples are selected per cycle, where $B$ is the number of selected samples, and $K$ represents the number of classes in each dataset. As demonstrated in \cite{bang2024active}, class-balanced sampling provides significant benefits. Therefore, we maintain a class-balanced setup in our active learning experiments for comparison with published baselines. We categorize the baselines into $O(n)$ methods, which require only a single pass over the unlabeled dataset (such as \textit{Entropy, Softmax, and Margin}), and $O(n^2)$ methods, such as \textit{BADGE and Coreset}, which involve feature extraction and distance computation on either labeled or unlabeled data. O-notation represents the Big-O notation and is used exclusively to measure the time complexity of the selection process.

\begin{figure*}[ht]
    \centering
    \begin{subfigure}[b]{0.24\textwidth} % 24% of text width to fit four in a row
        \centering
        \includegraphics[width=\textwidth]{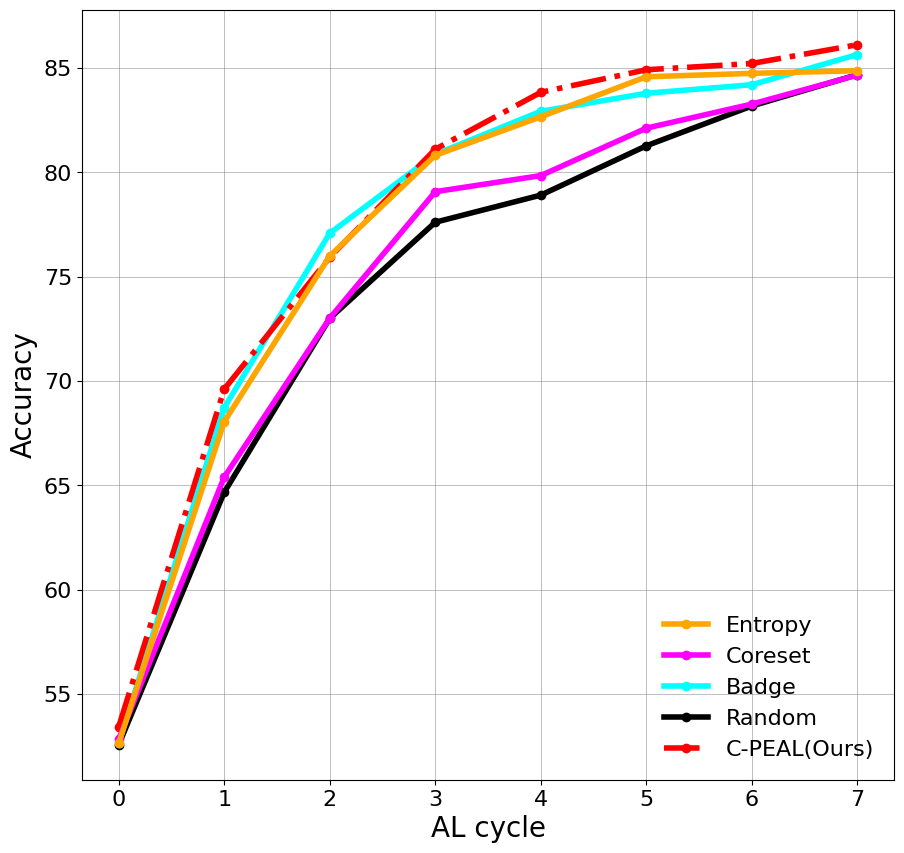} % Replace with actual image path
        \caption{Oxford Pets using ViT-B/16}
        \label{fig:OxfordPetsPL}
    \end{subfigure}
    \hfill
    \begin{subfigure}[b]{0.24\textwidth}
        \centering
        \includegraphics[width=\textwidth]{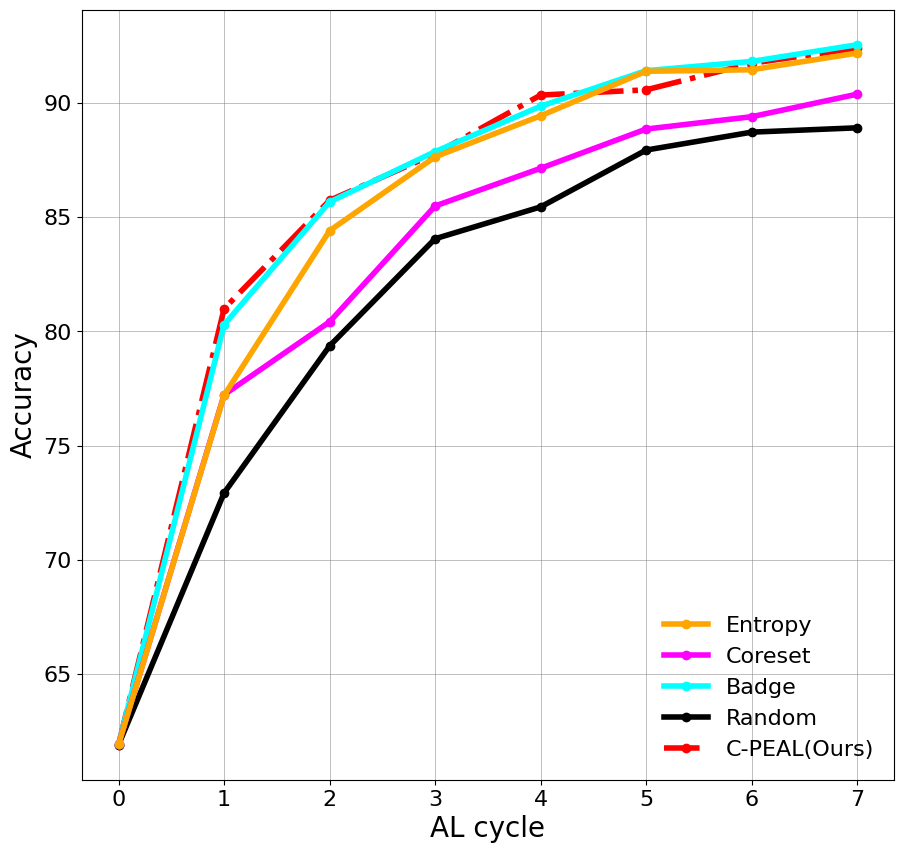}
        \caption{Caltech101 using ViT-B/16}
        \label{fig:Caltech101PL}
    \end{subfigure}
    \hfill
    \begin{subfigure}[b]{0.24\textwidth}
        \centering
        \includegraphics[width=\textwidth]{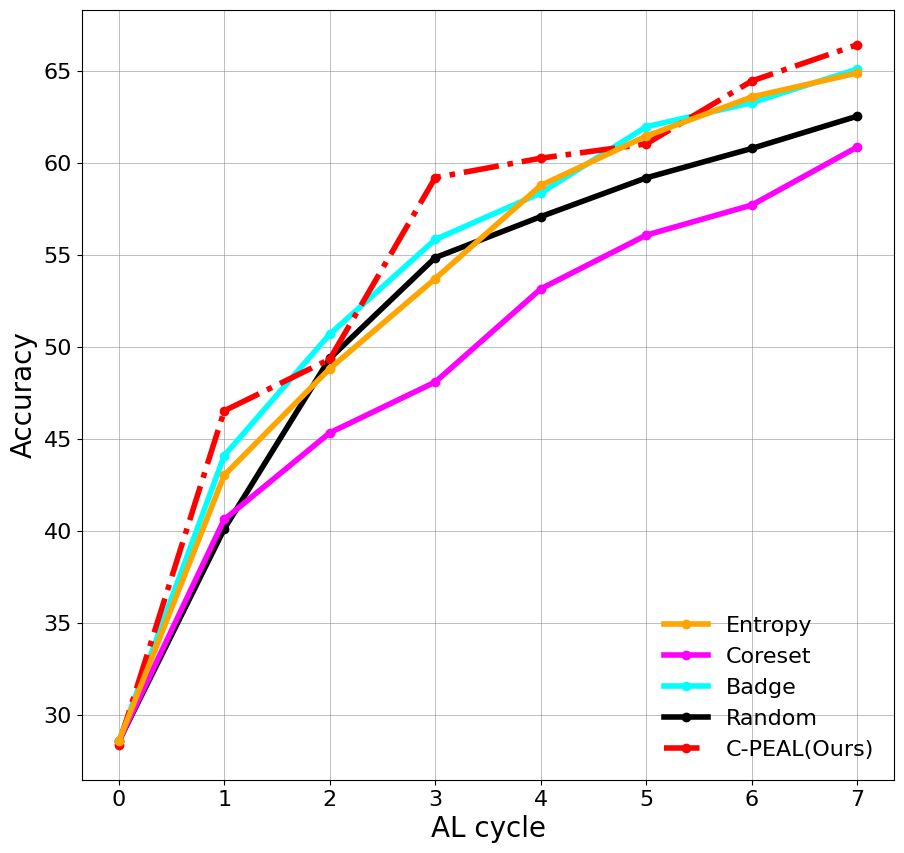}
        \caption{DTD using ViT-B/16}
        \label{fig:DTDPL}
    \end{subfigure}
    \hfill
    \begin{subfigure}[b]{0.24\textwidth}
        \centering
        \includegraphics[width=\textwidth]{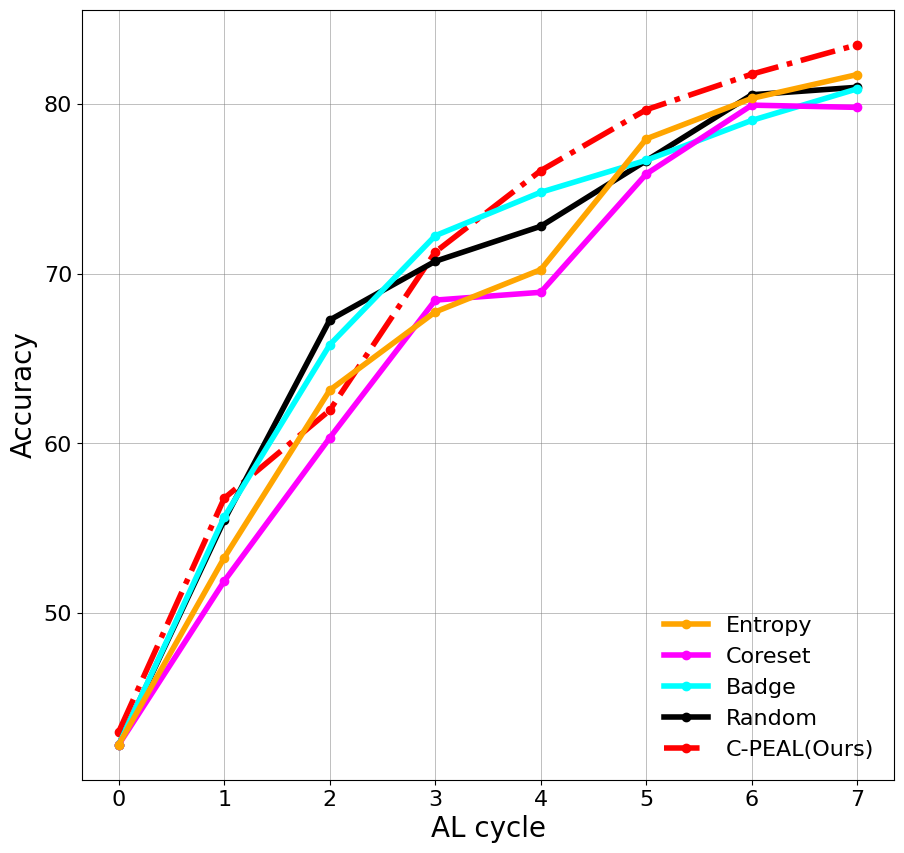}
        \caption{EuroSAT using ViT-B/16}
        \label{fig:EuroSATPL}
    \end{subfigure}
    
    \hfill
       \begin{subfigure}[b]{0.24\textwidth} % 24% of text width to fit four in a row
            \centering
            \includegraphics[width=\textwidth]{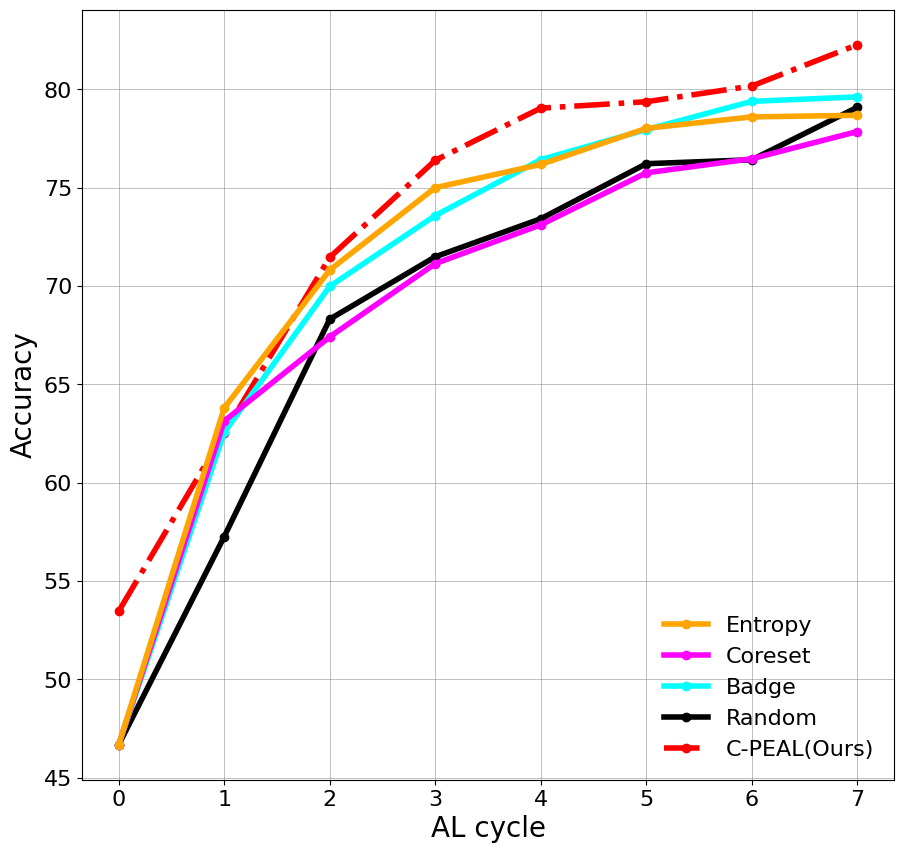} % Replace with actual image path
            \caption{Oxford Pets using ViT-B/32}
            \label{fig:OxfordPetsPL}
        \end{subfigure}
        \hfill
        \begin{subfigure}[b]{0.24\textwidth}
            \centering
            \includegraphics[width=\textwidth]{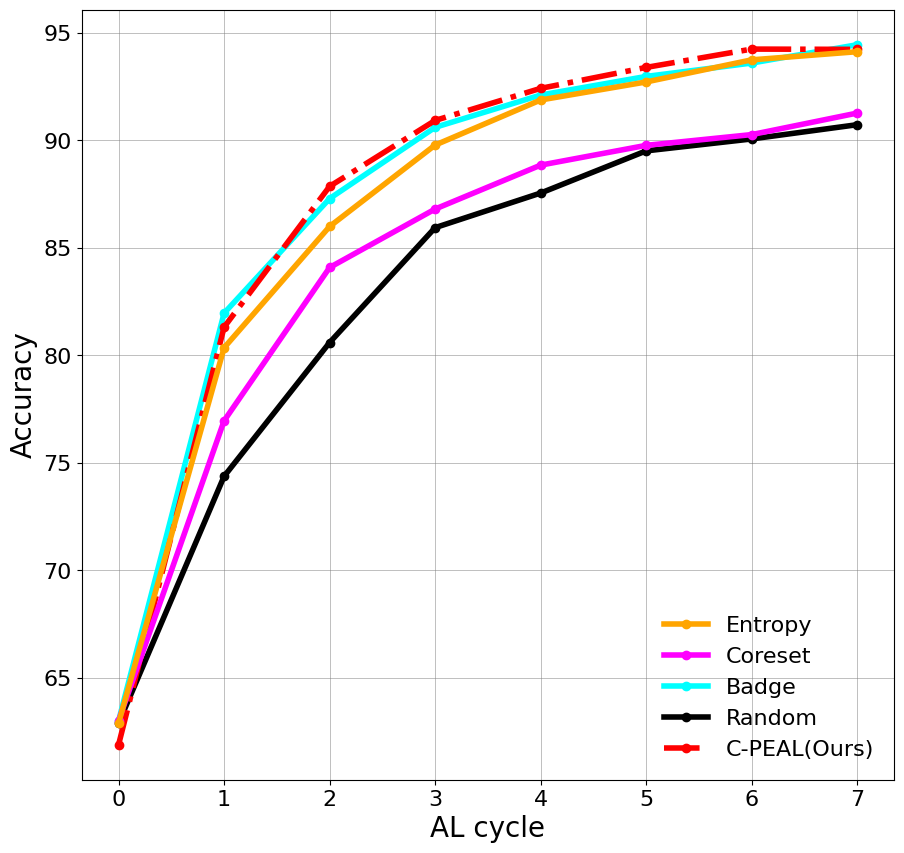}
            \caption{Caltech101 using ViT-B/32}
            \label{fig:Caltech101PL}
        \end{subfigure}
        \hfill
        \begin{subfigure}[b]{0.24\textwidth}
            \centering
            \includegraphics[width=\textwidth]{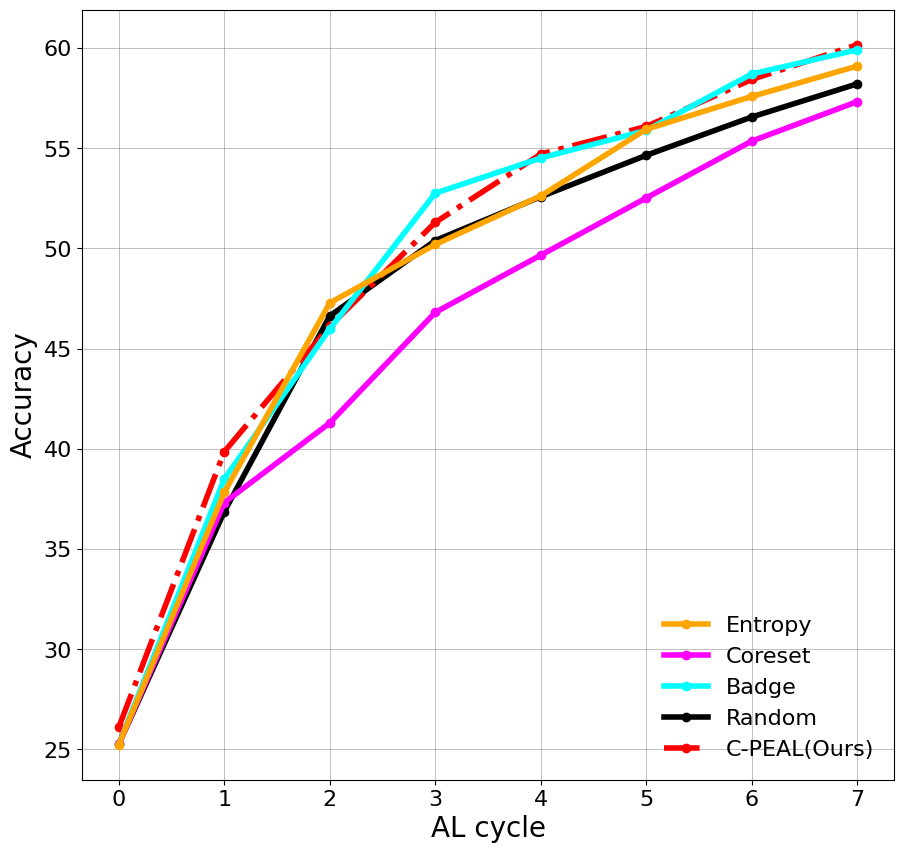}
            \caption{DTD using ViT-B/32}
            \label{fig:DTDPL}
        \end{subfigure}
        \hfill
        \begin{subfigure}[b]{0.24\textwidth}
            \centering
            \includegraphics[width=\textwidth]{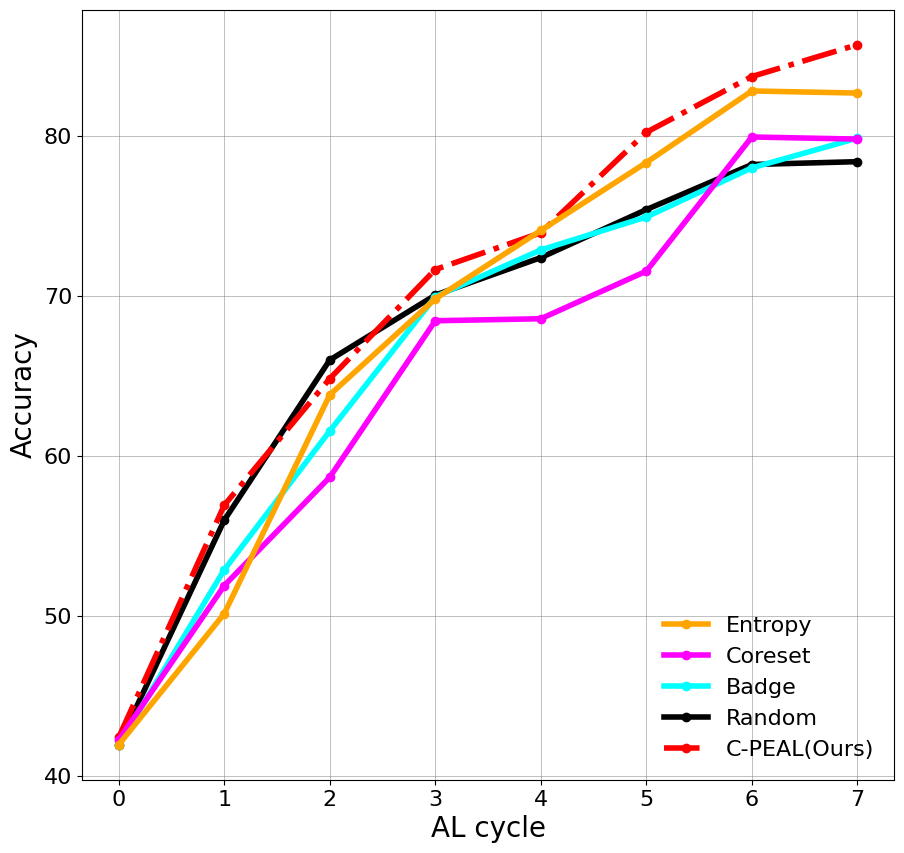}
            \caption{EuroSAT using ViT-B/32}
            \label{fig:EuroSATPL}
        \end{subfigure}    

    \hfill
       \begin{subfigure}[b]{0.24\textwidth} % 24% of text width to fit four in a row
            \centering
            \includegraphics[width=\textwidth]{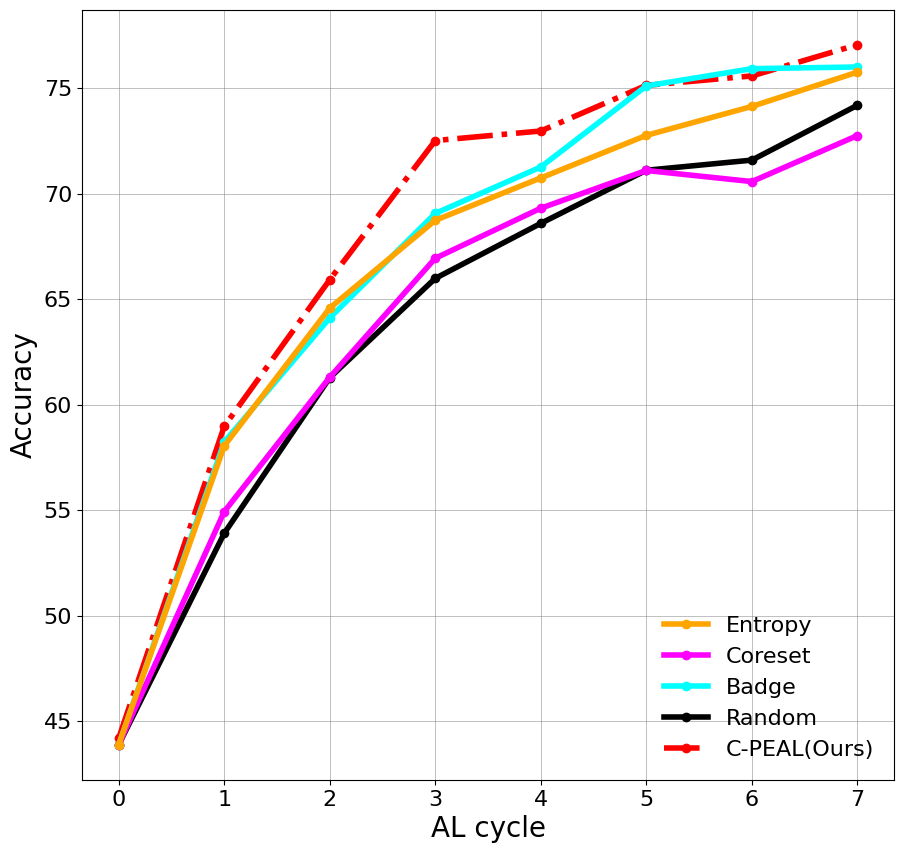} % Replace with actual image path
            \caption{Oxford Pets using R50}
            \label{fig:OxfordPetsPL}
        \end{subfigure}
        \hfill
        \begin{subfigure}[b]{0.24\textwidth}
            \centering
            \includegraphics[width=\textwidth]{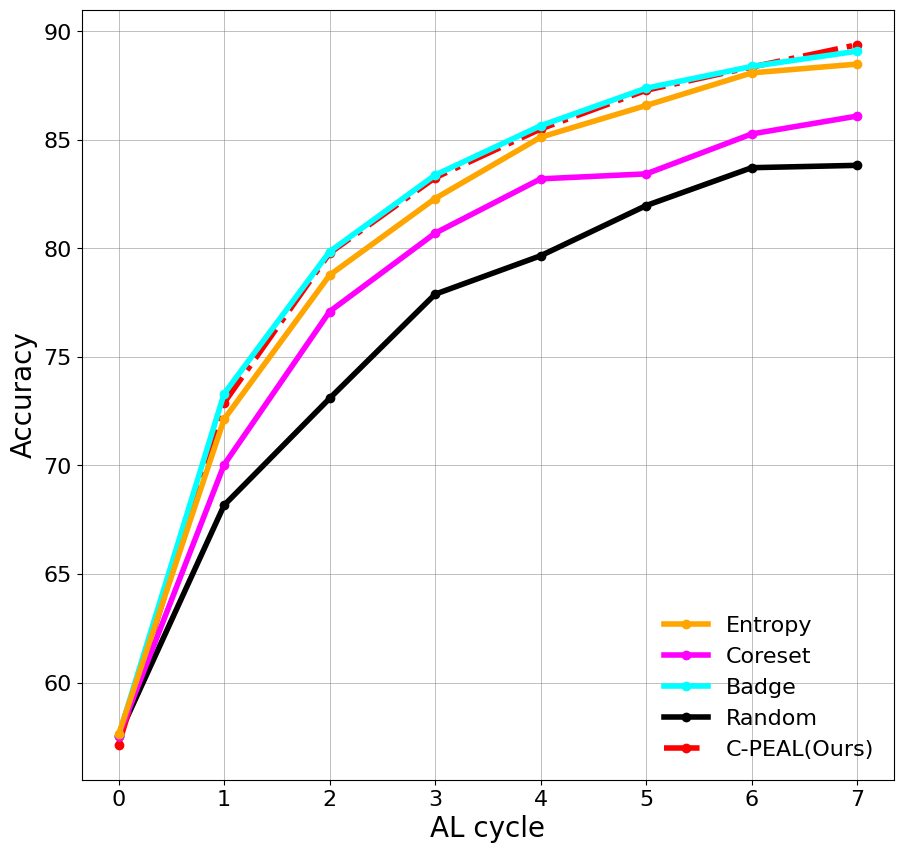}
            \caption{Caltech101 using R50}
            \label{fig:Caltech101PL}
        \end{subfigure}
        \hfill
        \begin{subfigure}[b]{0.24\textwidth}
            \centering
            \includegraphics[width=\textwidth]{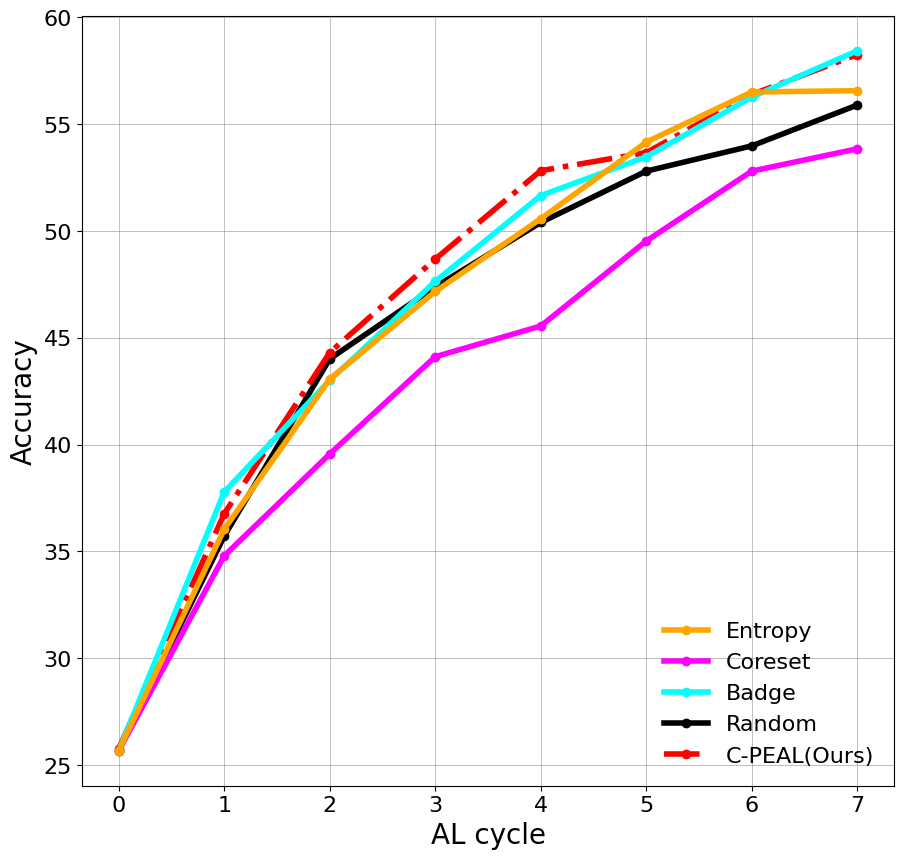}
            \caption{DTD using R50}
            \label{fig:DTDPL}
        \end{subfigure}
        \hfill
        \begin{subfigure}[b]{0.24\textwidth}
            \centering
            \includegraphics[width=\textwidth]{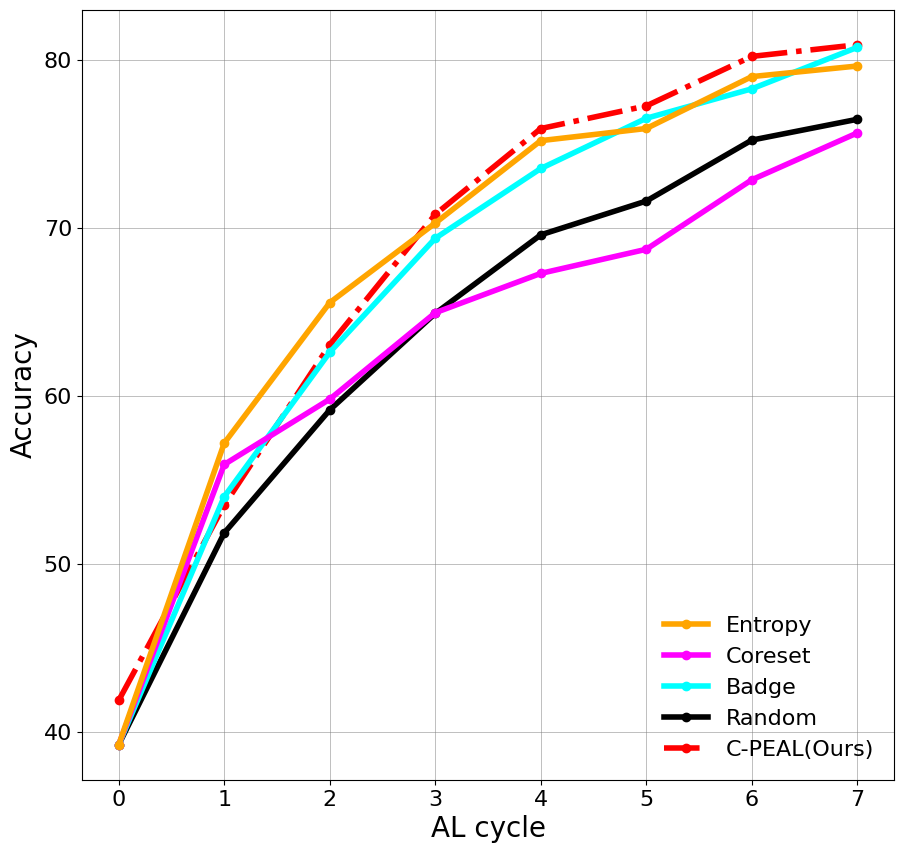}
            \caption{EuroSAT using R50}
            \label{fig:EuroSATPL}
        \end{subfigure}

    % \captionsetup{font=small}
    \caption{Active learning results for transformer models (ViT-B/16 and ViT-B/32) and ResNet-50 (R50) using the prompt learning method, evaluated per cycle. The mean of the three random seeds are shown in the plots.}
    \label{fig:promtresplots}
\end{figure*}

\subsection{Prompt learning Results}
We evaluate our calibrated learning loss methodology across three vision backbones and four datasets. Table \ref{table:prompt_learning_vit32_main} presents our results on the ViT-B/32 model, comparing it to zero-shot, random, and other baseline methods. Accuracy is used as the primary metric for comparison, and we report the average final accuracy over three seeds at the conclusion of the AL rounds. The variations of our methods are presented in the table, with each method described as follows: (1) the base method uses a fixed secondary loss weight, $\alpha$, determined through grid search; (2) \textit{+ANN} indicates that annealing is applied to the secondary loss weight; This approach enables the model to focus on optimizing critical tasks at different stages of training, such as prioritizing cross-entropy loss over calibration loss during the initial iterations for better convergence (3) \textit{+INTERW} employs computed balancing weights for incorrect and correct classifications ($\beta$ and $\gamma$) within the calibrated loss. Our proposed methods perform better than all the O($n$) uncertainty based sampling selection methods. Compared to Entropy \cite{bang2024active}, our method consistently outperforms across all datasets, achieving a final improvement of approximately $1$–$3$\%. This improvement is particularly notable in the EuroSAT dataset, with $+3.21$\% , considered as out-of-distribution (OOD), where challenges like poor zero-shot performance and class imbalance are present for CLIP \cite{DBLP:journals/corr/abs-2408-13320}. Given robust few-shot performance of CLIP models, even a 1\% improvement represents a significant advancement, underscoring the impact of calibration methods. In comparison to O($n^2$), our method outperforms approaches like BADGE across all datasets, though the gains are less pronounced for the DTD dataset. For the Oxford Pets dataset, while learning performs worse than zero-shot, as already evidenced in \cite{bang2024active}, our method narrows the gap more effectively than previous baselines. Sampling more data per cycle is one way to address this issue. 

To demonstrate the independence toward the vision backbone model, we present the best results per cycle for each dataset in Figure \ref{fig:promtresplots}, along with the impact of calibration in Table \ref{table:promptmodelplots}.
Overall, using balancing weights for the calibration loss proves to be beneficial irrespective of backbone model or dataset used in comparison to uncalibrated entropy. In AL, determining an appropriate stopping point can be challenging, as the labeling budget is often constrained by cost considerations. In Table \ref{table:promptmodelplots} we also showcase the results for cycle $4$ and cycle $8$ to signify that our method is beneficial at most or every cycle of AL.

\begin{table*}[h]
\centering
\resizebox{\textwidth}{!}{%
\begin{tabular}{cccccccccc}
\toprule
\multirow{2}{*}{Model} & \multirow{2}{*}{Method} & \multicolumn{2}{c}{Oxford Pets} & \multicolumn{2}{c}{DTD} & \multicolumn{2}{c}{Caltech101} & \multicolumn{2}{c}{EuroSAT} \\
\cmidrule(lr){3-4} \cmidrule(lr){5-6} \cmidrule(lr){7-8} \cmidrule(lr){9-10}
&& Cycle 4 & Cycle 8 & Cycle 4 & Cycle 8 & Cycle 4 & Cycle 8 & Cycle 4 & Cycle 8\\
\midrule
\multirow{4}{*}{R50} 
& Random & 65.99 ± 1.06 & 74.19 ± 1.08 & 47.38 ± 0.70 & 55.89 ± 0.82 & 79.66 ± 0.57 & 83.83 ± 0.53 & 64.91 ± 1.34 & 76.47 ± 2.08 \\
& Entropy & 68.74 ± 1.53 & 75.76 ± 1.00 & 47.16 ± 1.06 & 56.56 ± 1.00 & 85.13 ± 0.25 & 88.49 ± 0.15 & 70.29 ± 1.95 & 79.64 ± 1.29 \\
& BADGE & 69.07 ± 0.83 & 76.01 ± 1.32 & 47.64 ± 2.51 & \textbf{58.43 ± 0.73} & \textbf{85.65 ± 0.92} & 89.09 ± 0.26 & 69.38 ± 1.33 & 80.75 ± 0.89 \\
& C-PEAL +INTERW (ours) & \textbf{72.52 ± 1.60} & \textbf{77.06 ± 0.35} & \textbf{48.68 ± 0.56} & 58.25 ± 1.77 & 85.50 ± 1.07 & \textbf{89.38 ± 0.50} & \textbf{70.81 ± 2.16} &\textbf{ 80.91 ± 1.67} \\
\midrule
\multirow{4}{*}{ViT-B/16} 
& Random & 78.90 ± 2.34 & 84.67 ± 1.67 & 57.07 ± 0.37 & 62.53 ± 0.59 & 87.55 ± 0.78 & 90.72 ± 0.36 & 72.79 ± 3.62 & 80.98 ± 0.78 \\
& Entropy & 82.66 ± 1.94 & 84.86 ± 0.90 & 58.77 ± 1.29 & 64.87 ± 0.78 & 91.87 ± 0.45 & 94.12 ± 0.03 & 70.23 ± 7.08 & 81.73 ± 4.54 \\
& BADGE & 82.92 ± 1.46 & 85.64 ± 0.19 & 58.35 ± 0.71 & 65.09 ± 0.77 & 92.10 ± 0.16 & \textbf{94.44 ± 0.13} & 74.79 ± 2.28 & 80.89 ± 2.67 \\
& C-PEAL +INTERW (ours) & \textbf{83.83 ± 1.19} & \textbf{86.11 ± 0.98} & \textbf{60.25 ± 0.82} & \textbf{66.41 ± 0.42} & \textbf{92.41 ± 0.29} & 94.23 ± 0.28 & \textbf{76.07 ± 1.50} & \textbf{83.49 ± 0.96} \\
\midrule
% \bottomrule
\end{tabular}
}
% \captionsetup{font=small}
\caption{Performance comparison for different methods across models and datasets. Results are shown for AL Cycle 4 and Cycle 8. For completeness, all experiments have been rerun from \cite{bang2024active} with same seeds.}
\label{table:promptmodelplots}
\end{table*}

\subsection{Prompt vs LoRA}
When model internals are inaccessible, such as in closed models, Prompt learning is advantageous; however, LoRA offers a distinct benefit by allowing control over parameter training through rank selection, as it operates on a subset of parameters. To the best of our knowledge, this is the first AL study comparing these two approaches. For this comparison, we set the rank of low-rank matrices to \( r = 4 \) for Oxford Pets, DTD, and Caltech101, and \( r = 2 \) for EuroSAT, ensuring a comparable—or, in some cases, lower—number of trainable parameters than in prompt learning. This setup is especially advantageous for scenarios with fewer classes, the embedding size depends on the number of classes, which can lead to rapid growth in model size, as seen in Caltech101 dataset with $100$ classes. In contrast, LoRA allows trainable adapters to be flexibly inserted into both vision and text encoders at chosen layers.

% \begin{table}[h]
% \centering
% \small
% \begin{tabular}{lccc}
% \toprule
% \multirow{2}{*}{Dataset} & \multicolumn{2}{c}{Random} & \multirow{2}{*}{\begin{tabular}[c]{@{}c@{}}Trainable Parameters \\ (in millions)\end{tabular}} \\
% \cmidrule(lr){2-3}
%  & Cycle 1 & Cycle 8 \\
% \midrule
% Oxford Pets & +33.61 & +9.07 & +0.065 (r=4) \\
% DTD & +15.40 & +3.76 & -0.016 (r=4) \\
% Caltech101 & +27.53 & +4.87 & -0.450 (r=4) \\
% EuroSAT & +6.92 & +8.86 & +0.1024 (r=2) \\
% \bottomrule
% \end{tabular}
% \captionsetup{font=small}
% \caption{Comparison of performance improvements using Random selection with the LoRA method over Prompt learning across four datasets (Oxford Pets, DTD, Caltech101, and EuroSAT) at AL Cycles 1 and 8 with ViT-B/32. The adapter ranks (\textit{r}) used are specified. A \textquotedblleft +\textquotedblright\ indicates an increase in accuracy or parameters (in millions) for LoRA methods compared to Prompt learning, while a \textquotedblleft -\textquotedblright\ indicates a decrease. The last column shows the change in the number of parameters (in millions).}
% \label{table:promptvslora}
% \end{table}

\begin{table*}[h]
\centering
\fontsize{8}{10}\selectfont % Set font size to 8pt with a 10pt line spacing
\renewcommand{\arraystretch}{1.0} % Further reduce row height
\setlength{\tabcolsep}{3pt} % Reduce column padding
\begin{tabular}{ccccccc}
\toprule
 \textbf{Fine tuning method} & \textbf{Method} & \textbf{DTD} & \textbf{Oxford Pets} & \textbf{Caltech101} & \textbf{EuroSAT}  & \textbf{Avg Acc} \\ 
\midrule
% \\multirow{4}{*}{{Prompt learning} {6}{l}{Prompt learning} \\
% \midrule
\multirow{4}{*}{Prompt Learning \cite{lester2021power}}
&Random & 58.77 ± 1.94 & 78.30 ± 0.74 & 89.55 ± 1.00 & 77.62 ± 1.12 & 76.06 \\ 
&Entropy & 59.73 ± 1.96 & \underline{80.44 ± 1.24} & \underline{92.41 ± 0.50} & 80.80 ± 2.88 & 78.345 \\
&Coreset & 55.77 ± 1.33 & 76.84 ± 1.10 & 89.96 ± 0.03 & 77.50 ± 4.64  & 75.01\\ 
&BADGE & \textbf{60.28 ± 0.75} & 80.22 ± 1.69 & 92.21 ± 0.92 & \underline{81.98 ± 0.81} & \underline{78.67} \\ 
&C-PEAL (Ours) & \underline{60.05 ± 1.27} & \textbf{82.27 ± 0.17} & \textbf{94.10 ± 0.28} & \textbf{84.01 ± 0.12} & \textbf{80.10} \\ 
\midrule
% \midrule
% \multicolumn{6}{l}{LoRA} \\
\midrule
\multirow{4}{*}{LoRA \cite{hu2022lora}}
&Random & 61.98 ± 0.58 & 88.16 ± 0.46 & 93.78 ± 0.44  & 87.24 ± 1.90  & 82.78\\ 
&Entropy & 61.84 ± 1.50 & 88.37 ± 0.32  & 95.43 ± 0.45 &  90.44 ± 1.38 & 84.02\\
&Coreset & 60.78 ± 0.1 & 86.92 ± 0.10 & 94.12 ± 0.26 & 88.85 ± 0.92  & 82.67  \\ 
&BADGE & \underline{62.92 ± 0.87} & \underline{88.29 ±0.1} & \underline{95.69 ± 0.81} &  \underline{91.55 ± 0.66} & \underline{84.61} \\ 
&C-PEAL (Ours) & \textbf{63.94 ± 0.53} & \textbf{89.19 ± 0.96} & \textbf{96.23 ± 0.18} &   \textbf{91.74 ± 00.50}  & \textbf{85.27}\\ 
\bottomrule
\end{tabular}
% \captionsetup{font=small}
\caption{Final accuracy on four downstream tasks using the ViT-B/32 image encoder after eight rounds. Results compare Prompt-based and LoRA-based learning setups, with our method incorporating a weighting mechanism for calibration loss, achieving the highest performance across tasks. Best accuracies are marked in bold followed by underlined for second best.}
\label{table:resultspromptvslora}
\end{table*}

\begin{figure}
    \centering
    \includegraphics[width=0.67\linewidth]{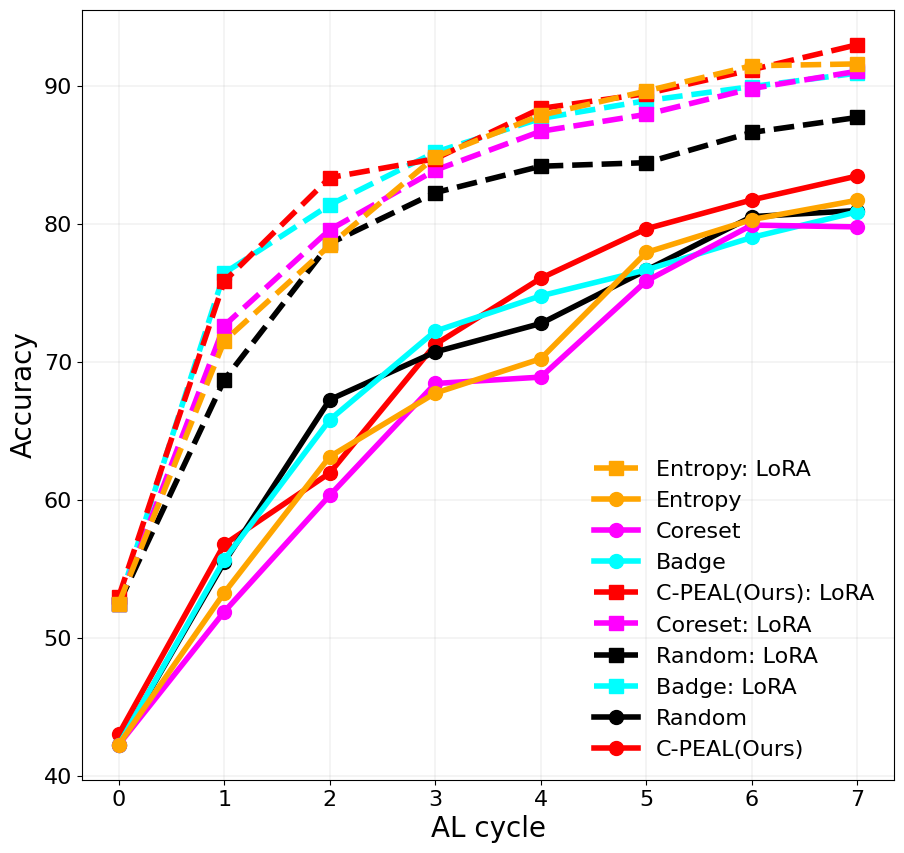}
    % \captionsetup{font=small}
    \caption{AL cycles for the EuroSAT dataset using ViT-B/16, comparing Prompt Learning and LoRA. Results demonstrate the performance differences across multiple cycles, underscoring the effectiveness of our approach irrespective of the fine-tuning method.}
    \label{fig:promptvslora}
\end{figure}

 We apply LoRA as outlined in \cite{zanella2024low} for ViT-B/16 and ViT-B/32 models. Surprisingly, LoRA yields a substantial performance boost and performs well even with random sampling, as shown in Figure \ref{fig:promptvslora}. LoRA achieves an average accuracy improvement of 20.86\% in Cycle 1 and 6.64\% in Cycle 8 over Prompt learning across four datasets using ViT-B/32, with the most significant gains observed in the early active learning cycles. Additionally, LoRA reduces the number of trainable parameters by an average of 0.075 million compared to Prompt learning, as its parameter count depends on the selected rank and number of layers rather than the number of classes. This performance gain, approaching zero-shot levels, highlights LoRA as a parameter-efficient fine-tuning option, particularly for datasets with fewer classes. This is a much more attractive solution for open source VLMs like CLIP where the model architecture is known. We also apply our enhanced sampling selection methods within a LoRA setup. A comparison between Prompt Tuning and LoRA is presented in Table \ref{table:resultspromptvslora} for ViT-B/32 model and Figure \ref{fig:promptvslora} for ViT-B/16.

\begin{figure}[h]
    \centering
    \begin{subfigure}[b]{0.49\columnwidth}
        \centering
        \includegraphics[width=\textwidth]{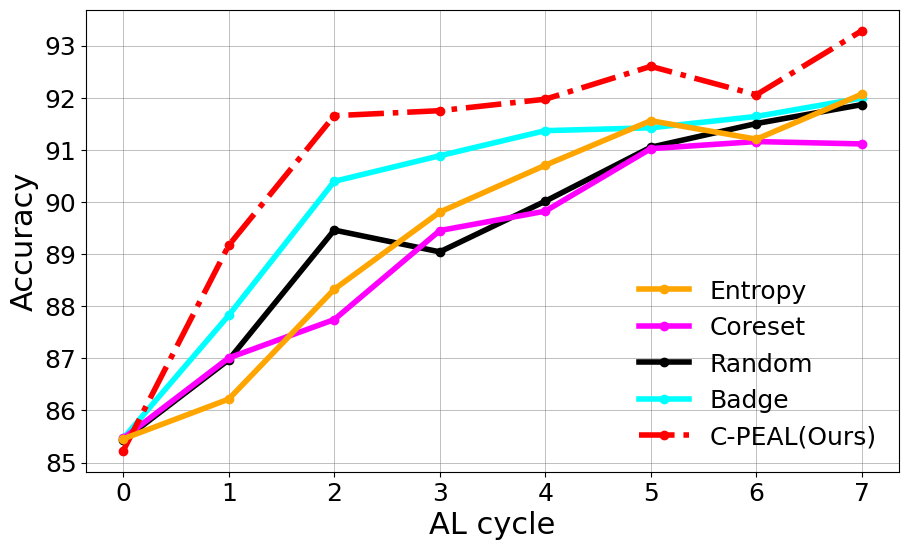}
        \caption{Oxford Pets}
        \label{fig:oxford_pets}
    \end{subfigure}
    \hfill
    \begin{subfigure}[b]{0.49\columnwidth}
        \centering
        \includegraphics[width=\textwidth]{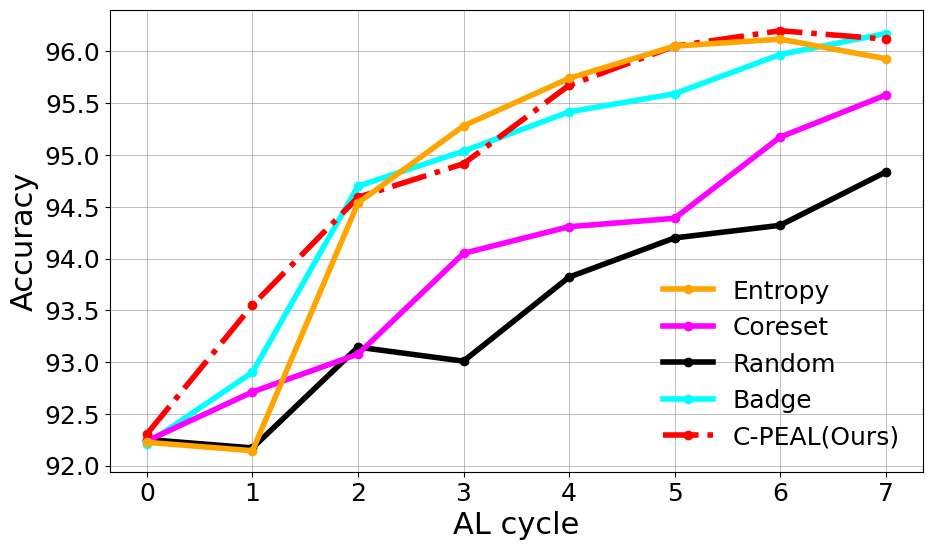}
        \caption{Caltech101}
        \label{fig:caltech101}
    \end{subfigure}
    \vspace{-3pt} % Reduces vertical space between rows
    \begin{subfigure}[b]{0.49\columnwidth}
        \centering
        \includegraphics[width=\textwidth]{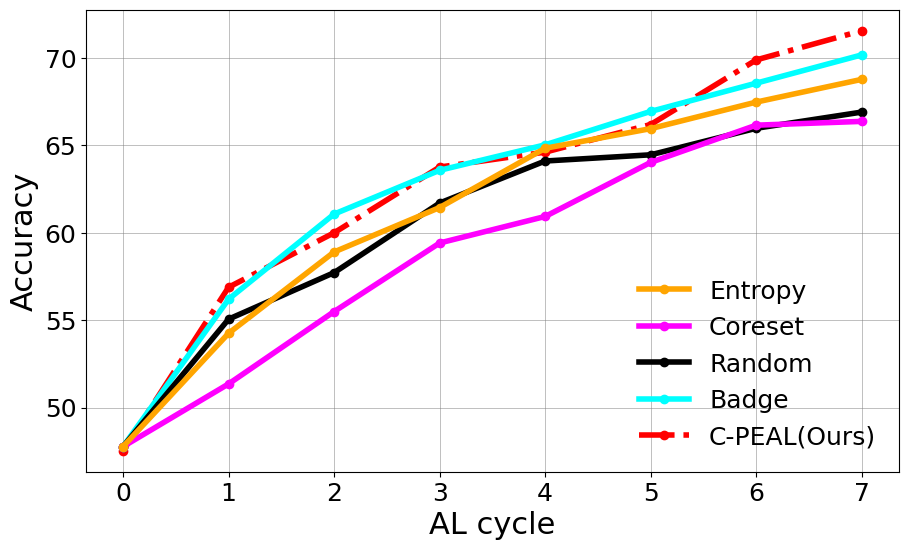}
        \caption{DTD}
        \label{fig:dtd}
    \end{subfigure}
    \hfill
    \begin{subfigure}[b]{0.49\columnwidth}
        \centering
        \includegraphics[width=\textwidth]{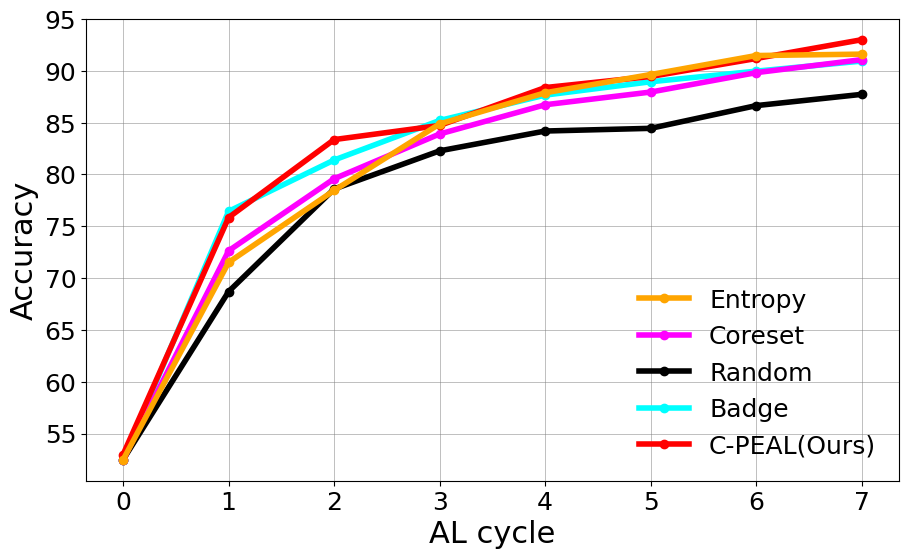}
        \caption{EuroSAT}
        \label{fig:eurosat}
    \end{subfigure}
    % \captionsetup{font=small}
    \caption{Model architecture of ViT-B/16 in a LoRA setup, integrating low-rank adapters into both text and vision encoders. The results highlight the advantages of our approach across various datasets: (a) Oxford Pets, (b) Caltech101, (c) DTD, and (d) EuroSAT. }
    \label{fig:lora_res_vut16}
\end{figure}

\subsection{LoRA Results}
We integrate LoRA adapters into the transformer-based vision backbones for both ViT-B/16 and ViT-B/32, as well as the text encoder. Using LoRA, we observe not only improved sampling quality but also the highest performance across all datasets as shown in Table \ref{table:resultspromptvslora}. Notably, CLIP surpasses zero-shot performance comfortably for each dataset. Specifically, we achieve improvements of +3.89\% for DTD, +6.92\% for Oxford Pets, +2.13\% for Caltech101, and +7.51\% for EuroSAT. Overall, the average accuracy across all datasets with the ViT-B/32 model increases by 5.17\%.
% \begin{table*}[h]
% \centering
% \fontsize{8}{10}\selectfont % Set font size to 8pt with a 10pt line spacing
% \renewcommand{\arraystretch}{1.0} % Further reduce row height
% \setlength{\tabcolsep}{3pt} % Reduce column padding
% \begin{tabular}{lcccccccc}
% \toprule
% \multirow{2}{*}{\textbf{Method}} & \multicolumn{4}{c}{\textbf{Prompt}} & \multicolumn{4}{c}{\textbf{LoRA}} \\
% \cmidrule(lr){2-5} \cmidrule(lr){6-9}
%  & \textbf{DTD} & \textbf{Oxford Pets} & \textbf{Caltech101} & \textbf{Eurosat} & \textbf{DTD} & \textbf{Oxford Pets} & \textbf{Caltech101} & \textbf{Eurosat} \\ 
% \midrule
% Random & 61.98 ± 0.58 & 88.16 ± 0.46 & 93.78 ± 0.44 & 00.00 ± 00.00 & 62.10 ± 0.60 & 88.30 ± 0.45 & 94.00 ± 0.42 & 00.00 ± 00.00 \\ 
% Entropy & 61.84 ± 1.50 & 88.37 ± 0.32 & 95.43 ± 0.45 & 00.00 ± 00.00 & 62.05 ± 1.45 & 88.50 ± 0.30 & 95.60 ± 0.43 & 00.00 ± 00.00 \\
% BADGE & \underline{62.92 ± 0.87} & \underline{88.29 ±0.1} & \underline{95.69 ± 0.81} & 00.00 ± 00.00 & 63.00 ± 0.85 & 88.50 ± 0.10 & 95.80 ± 0.80 & 00.00 ± 00.00 \\ 
% \midrule
% C-PEAL \\
% + INTERW & \textbf{63.94 ± 0.53} & \textbf{89.19 ± 0.96} & \textbf{96.23 ± 0.18} & 00.00 ± 00.00 & \textbf{64.00 ± 0.50} & \textbf{89.30 ± 0.90} & \textbf{96.40 ± 0.20} & 00.00 ± 00.00 \\ 
% \bottomrule
% \end{tabular}
% \captionsetup{font=small}
% \caption{Final accuracy on four downstream tasks with ViT-B/32 image encoder. Final accuracy is the accuracy after eight rounds for both prompt-based and LoRA-based methods}
% \label{table:prompt_vs_lora_comparison}
% \end{table*}

For the ViT-B/16 model, we present the cycle-wise AL results in Figure \ref{fig:lora_res_vut16}. Our model performs best on almost all the datasets and showcases gains at earlier AL cycles. Using our method, we observe overall improvements over Entropy-based sampling of +2.79\% for DTD, +1.01\% for Caltech101, +1.13\% for EuroSAT, and +1.39\% for Oxford Pets. However for Caltech101, we slightly underperform compared to BADGE by only -0.15\%, while outperforming on all other datasets.

\subsection{Runtime Efficiency and Optimization}
The uncertainty calibration loss technique is runtime-efficient, as it adds only a loss component to the primary cross-entropy loss. This approach incurs minimal computational overhead while enhancing calibration and accuracy. As shown in Table \ref{table:relative_time}, the runtime aligns with \(O(n)\) methods. Overall, the proposed solution is optimal in both training parameters and sampling strategy, making it an attractive parameter-efficient sampling approach.

\section{Conclusion}
\label{sec:conclusion}
This paper addresses AL in a few-shot setting for vision-language models, presenting a novel parameter-efficient uncertainty calibration approach in active learning. Our method outperforms existing feature-based and uncertainty-based sampling methods while being computationally efficient. The proposed calibration loss is modular, adaptable to diverse vision backbones and foundation models. We hope this work will benefit the AI and neural network community by advancing data-efficient and compute-efficient adaptation of vision-language foundational models to various domains.
%Given the gains achieved by LoRA in AL, we plan to expand and benchmark additional PEFT methods across tasks, including high-cost annotation challenges like visual question answering.

% \begin{thebibliography}{00}
% \bibitem{b1} G. Eason, B. Noble, and I. N. Sneddon, ``On certain integrals of Lipschitz-Hankel type involving products of Bessel functions,'' Phil. Trans. Roy. Soc. London, vol. A247, pp. 529--551, April 1955.
% \bibitem{b2} J. Clerk Maxwell, A Treatise on Electricity and Magnetism, 3rd ed., vol. 2. Oxford: Clarendon, 1892, pp.68--73.
% \bibitem{b3} I. S. Jacobs and C. P. Bean, ``Fine particles, thin films and exchange anisotropy,'' in Magnetism, vol. III, G. T. Rado and H. Suhl, Eds. New York: Academic, 1963, pp. 271--350.
% \bibitem{b4} K. Elissa, ``Title of paper if known,'' unpublished.
% \bibitem{b5} R. Nicole, ``Title of paper with only first word capitalized,'' J. Name Stand. Abbrev., in press.
% \bibitem{b6} Y. Yorozu, M. Hirano, K. Oka, and Y. Tagawa, ``Electron spectroscopy studies on magneto-optical media and plastic substrate interface,'' IEEE Transl. J. Magn. Japan, vol. 2, pp. 740--741, August 1987 [Digests 9th Annual Conf. Magnetics Japan, p. 301, 1982].
% \bibitem{b7} M. Young, The Technical Writer's Handbook. Mill Valley, CA: University Science, 1989.
\bibliographystyle{IEEEtran}
\bibliography{IEEEabrv,conference_10719}
% \end{thebibliography}

\vspace{12pt}
\end{document}